%% file: proceedings.tex
\documentclass[10pt,twocolumn,letterpaper]{article}

\usepackage{cvpr}
\usepackage{times}
\usepackage{epsfig}
\usepackage{graphicx}
\usepackage{amsmath}
\usepackage{amssymb}

\usepackage{balance}       \usepackage{subfig}
\usepackage[T1]{fontenc}   \usepackage{txfonts}
\usepackage{mathptmx}
\usepackage{color}
\usepackage{booktabs}
\usepackage{textcomp}
\usepackage{cuted}
\usepackage{capt-of}

\usepackage{gensymb}

\usepackage[pagebackref=true,breaklinks=true,letterpaper=true,colorlinks,bookmarks=false]{hyperref}

\cvprfinalcopy 

\def\cvprPaperID{6270} 

\begin{document}

\title{Intuitive, Interactive Beard and Hair Synthesis with Generative Models}

\author{Kyle Olszewski$^1$$^3$, Duygu Ceylan$^2$, Jun Xing$^3$$^5$, Jose Echevarria$^2$, Zhili Chen$^2$$^7$, Weikai Chen$^3$$^6$, and Hao Li$^1$$^3$$^4$\\
  $^1$University of Southern California, $^2$Adobe Inc., $^3$USC ICT, $^4$Pinscreen Inc.,
  \\$^5$miHoYo, $^6$Tencent America, $^7$ByteDance Research  \\
  {\tt\small \{olszewski.kyle,duygu.ceylan,junxnui\}@gmail.com echevarr@adobe.com} \\
  {\tt\small \{iamchenzhili,chenwk891\}@gmail.com hao@hao-li.com}
}

\maketitle

\begin{abstract}
\input{uist_submission/abstract}
\end{abstract}

\setlength{\belowcaptionskip}{-5pt}

\section{Introduction}
\input{uist_submission/introduction}

\section{Related Work}
\input{uist_submission/relatedWork}
\label{sec:prior}

\section{Overview}
\input{uist_submission/overview}

\label{sec:overview}

\section{Network Architecture and Training}
\input{uist_submission/network}

\section{Dataset}
\input{uist_submission/dataset}
\label{sec:dataset}

\section{Interactive Editing}
\label{sec:ui}
\input{uist_submission/ui_cvpr.tex}

\section{Results}
\label{sec:results}
\input{uist_submission/results.tex}

\begin{figure}[h!]
\vspace{-2mm}
  \includegraphics[width=1.0\columnwidth]{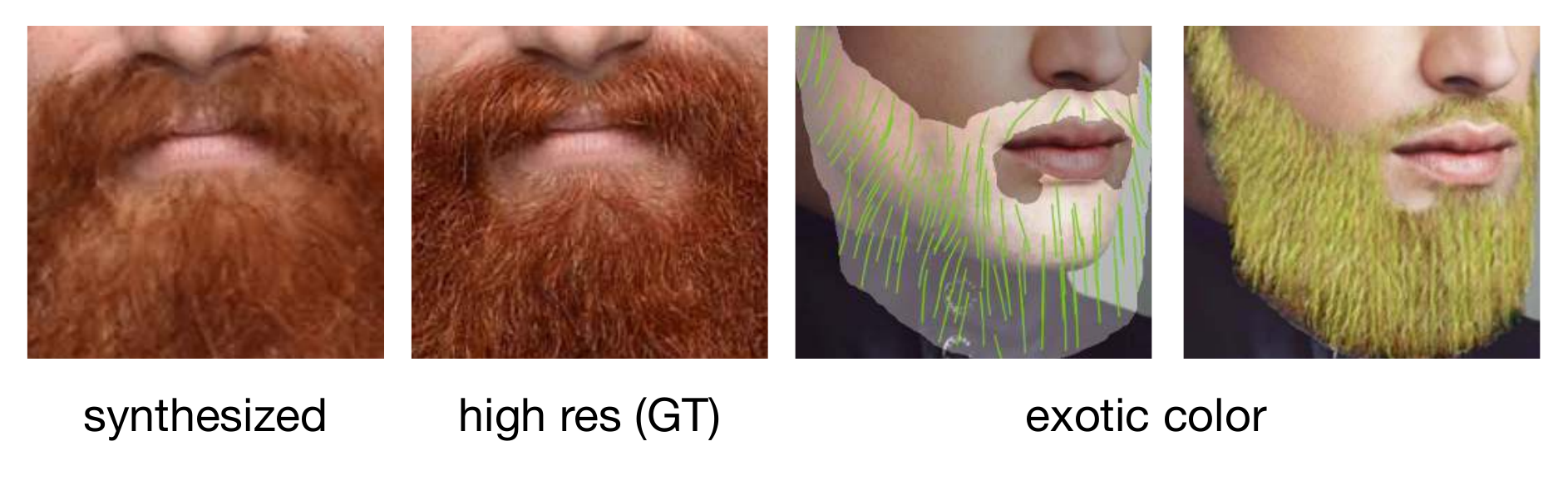}
  \vspace{-8mm}
  \caption{Limitations: our method does not produce satisfactory results in some extremely challenging cases.}
  \label{fig:limitation}
  \vspace{-5mm}
\end{figure}

\section{Limitations and Future Work}
\label{sec:conclusion}

\input{uist_submission/conclusion.tex}

{\small
\bibliographystyle{ieee_fullname}
\bibliography{paper}
}

\clearpage

\renewcommand{\thesection}{A\arabic{section}}
\setcounter{section}{0}

\input{uist_submission/supplementary.tex}

\end{document}

%% file: uist_submission/abstract.tex
We present an interactive approach to synthesizing realistic variations in facial hair in images, ranging from subtle edits to existing hair to the addition of complex and challenging hair in images of clean-shaven subjects. To circumvent the tedious and computationally expensive tasks of modeling, rendering and compositing the 3D geometry of the target hairstyle using the traditional graphics pipeline, we employ a neural network pipeline that synthesizes realistic and detailed images of facial hair directly in the target image in under one second. The synthesis is controlled by simple and sparse guide strokes from the user defining the general structural and color properties of the target hairstyle. We qualitatively and quantitatively evaluate our chosen method compared to several alternative approaches. We show compelling interactive editing results with a prototype user interface that allows novice users to progressively refine the generated image to match their desired hairstyle, and demonstrate that our approach also allows for flexible and high-fidelity scalp hair synthesis.

%% file: uist_submission/introduction.tex
The ability to create and edit realistic facial hair in images has several important, wide-ranging applications. For example, law enforcement agencies could provide multiple images portraying how missing or wanted individuals would look if they tried to disguise their identity by growing a beard or mustache, or how such features would change over time as the subject aged. Someone considering growing or changing their current facial hair may want to pre-visualize their appearance with a variety of potential styles without making long-lasting changes to their physical appearance. Editing facial hair in pre-existing images would also allow users to enhance their appearance, for example in images used for their social media profile pictures. Insights into how to perform high-quality and controllable facial hair synthesis would also prove useful in improving face-swapping technology such as Deepfakes for subjects with complex facial hair.

\begin{figure}[t!]
\includegraphics[width=\columnwidth]{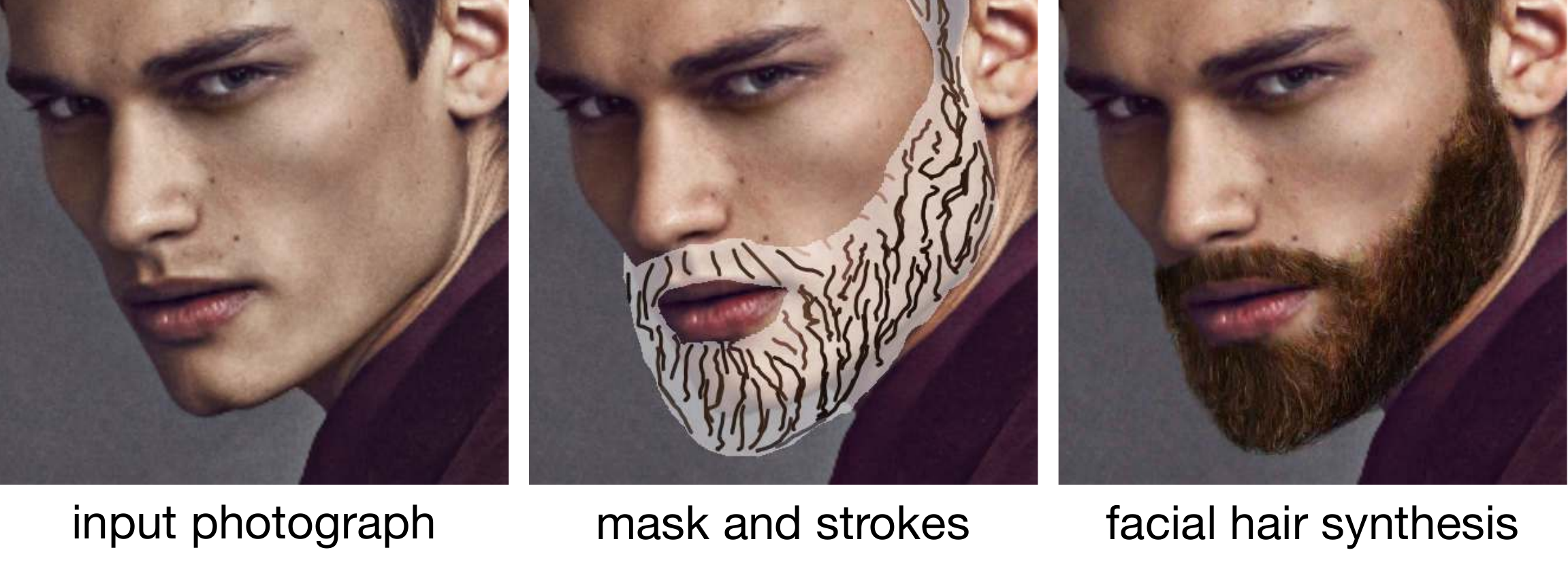}
\caption{Given a target subject image, a masked region in which to perform synthesis, and a set of strokes of varying colors provided by the user, our approach interactively synthesizes hair with the appropriate structure and appearance. \label{fig:teaser}}
\vspace{-7mm}
\end{figure}

One approach would be to infer the 3D geometry and appearance of any facial hair present in the input image, then manipulate or replace it as as desired before rendering and compositing into the original image.
However, single view 3D facial reconstruction is in itself an ill-posed and under constrained problem, and most state-of-the-art approaches struggle in the presence of large facial hair, and rely on parametric facial models which cannot accurately represent such structures.
Furthermore, even state-of-the-art 3D hair rendering methods would struggle to provide sufficiently realistic results quickly enough to allow for interactive feedback for users exploring numerous subtle stylistic variations.

One could instead adopt a more direct and naive approach, such as copying regions of facial hair from exemplar images of a desired style into the target image. However, it would be extremely time-consuming and tedious to either find appropriate exemplars matching the position, perspective, color, and lighting conditions in the target image, or to modify these properties in the selected exemplar regions so as to assemble them into a coherent style matching both the target image and the desired hairstyle.

In this paper we propose a learning-based interactive approach to image-based hair editing and synthesis.
We exploit the power of generative adversarial networks (GANs), which have shown impressive results for various image editing tasks.
However, a crucial choice in our task is the input to the network guiding the synthesis process used during training and inference.
This input must be sufficiently detailed to allow for synthesizing an image that corresponds to the user's desires.
Furthermore, it must also be tractable to obtain training data and extract input closely corresponding to that provided by users, so as to allow for training a generative model to perform this task.
Finally, to allow novice artists to use such a system, authoring this input should be intuitive, while retaining interactive performance to allow for iterative refinement based on realtime feedback.

A set of sketch-like ``guide strokes'' describing the local shape and color of the hair to be synthesized is a natural way to represent such input that corresponds to how humans draw images. Using straightforward techniques such as edge detection or image gradients would be an intuitive approach to automatically extract such input from training images.
However, while these could roughly approximate the types of strokes that a user might provide when drawing hair, we seek to find a representation that lends itself to intuitively editing the synthesis results without explicitly erasing and replacing each individual stroke. 

Consider a vector field defining the dominant local orientation across the region in which hair editing and synthesis is to be performed.
This is a natural representation for complex structures such as hair, which generally consists of strands or wisps of hair with local coherence, which could easily be converted to a set of guide strokes by integrating the vector field starting from randomly sampled positions in the input image.
However, this representation provides additional benefits that enable more intuitive user interaction.
By extracting this vector field from the original facial hair in the region to be edited, or by creating one using a small number of coarse brush strokes, we could generate a dense set of guide strokes from this vector field that could serve as input to the network for image synthesis. Editing this vector field would allow for adjusting the overall structure of the selected hairstyle (\eg, making a straight hairstyle more curly or tangled, or vice versa) with relatively little user input, while still synthesizing a large number of guide strokes corresponding to the user's input. As these strokes are used as the final input to the image synthesis networks, subtle local changes to the shape and color of the final image can be accomplished by simply editing, adding or removing individual strokes.

We carefully chose our network architectures and training techniques to allow for high-fidelity image synthesis, tractable training with appropriate input data, and interactive performance.
Specifically, we propose a two-stage pipeline. While the first stage focuses on synthesizing realistic facial hair, the second stage aims to refine this initial result and generate plausible compositions of the generated hair within the input image. 

The success of such a learning-based method depends on the availability of a large-scale training set that covers a wide range of facial hairstyles. To our knowledge, no such dataset exists, so we fill this void by creating a new synthetic dataset that provides variation along many axes such as the style, color, and viewpoint in a controlled manner. We also collect a smaller dataset of real facial hair images we use to allow our method to better generalize to real images. We demonstrate how our networks can be trained using these datasets to achieve realistic results despite the relatively small amount of real images used during training.

We introduce a user interface with tools that allow for intuitive creation and manipulation of the vector fields used to generate the input to our synthesis framework.
We conduct comparisons to alternative approaches, as well as extensive ablations demonstrating the utility of each component of our approach. Finally, we perform a perceptual study to evaluate the realism of images authored using our approach, and a user study to evaluate the utility of our proposed user interface. These results demonstrate that our approach is indeed a powerful and intuitive approach to quickly author realistic illustrations of complex hairstyles.

%% file: uist_submission/relatedWork.tex
\paragraph{Texture Synthesis} 
As a complete review of example-based texture synthesis methods is out of the scope of this paper, we refer the reader to the surveys of \cite{wei2009state, barnes2017survey} for comprehensive overviews of modern texture synthesis techniques.
In terms of methodology, example-based texture synthesis approaches can be mainly categorized into pixel-based methods \cite{Wei:2000:FTS, Efros:1999:TSN}, stitching-based methods \cite{Efros:2001:IQT, Kwatra:2003:GTI,lasram2012parallel},  optimization-based approaches \cite{Kwatra:2005:TOE,han2006fast,4069262,kaspar2015self} and appearance-space texture synthesis \cite{lefebvre2006appearance}. 
Close to our work, Luk{\'a}{\v c}~\etal~\cite{Lukac:2013:PFT} present a method that allows users to paint using the visual style of an arbitrary example texture.
In \cite{Lukac:2015:BEE}, an intuitive editing tool is developed to support example-based painting that globally follows user-specified shapes while generating interior content that preserves the textural details of the source image. This tool is not specifically designed for hair synthesis, however, and thus lacks local controls that users desire, as shown by our user study.

Recently, many researchers have attempted to leverage neural networks for texture synthesis~\cite{Gatys:2015:TSU,Li:2016:PRT,olszewski2017realistic}. 
However, it remains nontrivial for such techniques to accomplish simple editing operations, \eg changing the local color or structure of the output, which are necessary in our scenario.

\paragraph{Style Transfer} 
The recent surge of style transfer research suggests an alternate approach to replicating stylized features from an example image to a target domain~\cite{Gatys:2015:ANA,Johnson:2016:PLR,Liao:2017:VAT,saito2016photorealistic}. 
However, such techniques make it possible to handle varying styles from only one exemplar image. 
When considering multiple frames of images, a number of works have been proposed to extend the original technique to handle video \cite{RuderDB2016,selim2016painting} and facial animations \cite{Fiser:2017:ESS}.
Despite the great success of such neural-based style transfer techniques, one key limitation lies in their inability to capture fine-scale texture details.
Fi{\v{s}}er~\etal~\cite{fivser2016stylit} present a non-parametric model that is able to reproduce such details. 
However, the guidance channels employed in their approach is specially tailored for stylized 3D rendering, limiting its application.

\paragraph{Hair Modeling}
Hair is a crucial component for photorealistic avatars and CG characters.
In professional production, human hair is modeled and rendered with sophisticated devices and tools \cite{choe2005statistical,kim2002interactive,weng2013hair,yuksel2009hair}. 
We refer to \cite{ward2007survey} for an extensive survey of hair modeling techniques. In recent years, several multi-view~\cite{luo2012multi,hu2014robust} and single-view~\cite{chai2012single,Chai:2013:DHM,Hu:2015:SHM,chai2016autohair} hair modeling methods have been proposed.
An automatic pipeline for creating a full head avatar from a single portrait image has also been proposed \cite{hu2017avatar}.
Despite the large body of work in hair modeling, however, techniques applicable to facial hair reconstruction remain largely unexplored.
In \cite{beeler2012coupled}, a coupled 3D reconstruction method is proposed to recover both the geometry of sparse facial hair and its underlying skin surface. More recently, Hairbrush \cite{Xing:2019:HID} demonstrates an immersive data-driven modeling system for 3D strip-based hair and beard models.

\begin{figure*}[t!]
\centering
\includegraphics[width=0.85\textwidth]{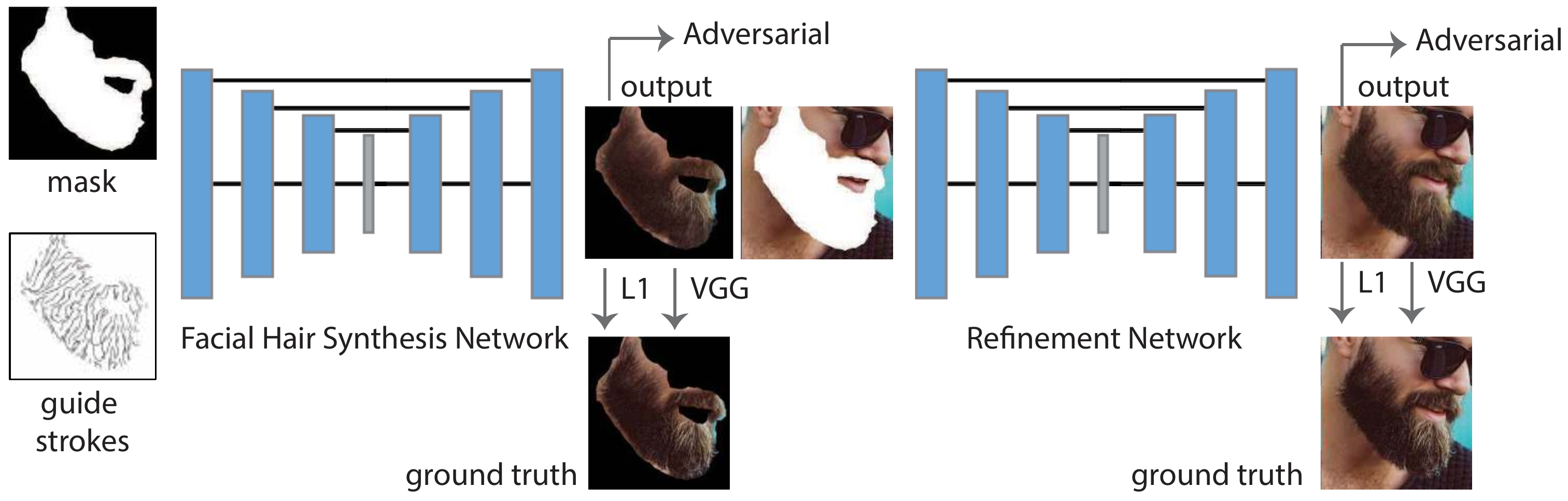}
    \vspace{-2mm}
  \caption{We propose a two-stage network architecture to synthesize realistic facial hair. Given an input image with a user-provided region of interest and sparse guide strokes defining the local color and structure of the desired hairstyle, the first stage synthesizes the hair in this region. The second stage refines and composites the synthesized hair into the input image.}
  \label{fig:pipeline}
  \vspace{-3mm}
\end{figure*}

\paragraph{Image Editing}
Interactive image editing has been extensively explored in computer graphics community over the past decades. Here, we only discuss prior works that are highly related to ours.
In the seminal work of Bertalmio~\etal~\cite{bertalmio2000image}, a novel technique is introduced to digitally inpaint missing regions using isophote lines.
P{\'e}rez~\etal~\cite{perez2003poisson} later propose a landmark algorithm that supports general interpolation machinery by solving Poisson equations.
Patch-based approaches \cite{Efros:2001:IQT,bertalmio2003simultaneous, criminisi2004region, barnes2009patchmatch, darabi2012image} provide a popular alternative solution by using image patches adjacent to missing context or in a dedicated source image to replace the missing regions.
Recently, several techniques \cite{isola2016image,sangkloy2017scribbler,zhu2017unpaired,park2019SPADE} based on deep learning have been proposed to translate the content of a given input image to a target domain.

Closer to our work, a number of works investigate editing techniques that directly operate on semantic image attributes.
Nguyen~\etal~\cite{nguyen2008image} propose to edit and synthesize beards by modeling faces as a composition of multiple layers.
Mohammed~\etal~\cite{mohammed2009visio} perform facial image editing by leveraging a parametric model learned from a large facial image database.
Kemelmacher-Shlizerman~\cite{kemelmacher2016transfiguring} presents a system that enables editing the visual appearance of a target portrait photo by replicating the visual appearance from a reference image.
Inspired by recent advances in deep neural networks, Brock~\etal~\cite{brock2016neural} propose a neural algorithm to make large semantic changes to natural images.
This technique has inspired follow-up works which leverage deep generative networks for eye inpainting \cite{dolhansky2017eye}, semantic feature interpolation \cite{upchurch2016deep} and face completion \cite{yeh2017semantic}.
The advent of generative adversarial networks (GANs) \cite{goodfellow2014generative} has inspired a large body of high-quality image synthesis and editing approaches \cite{chang2018pairedcyclegan,zhang2017stackgan,zhu2016generative,denton2015deep,shrivastava2017learning} using the power of GANs to synthesize complex and realistic images.
The latest advances in sketch \cite{Portenier:2018, DBLP:journals/corr/abs-1902-06838} or contour \cite{dekel2018sparse} based facial image editing enables users to manipulate facial features via intuitive sketching interfaces or copy-pasting from exemplar images while synthesizing results plausibly corresponding to the provided input. While our system also uses guide strokes for hair editing and synthesis, we find that intuitively synthesizing realistic and varied facial hair details requires more precise control and a training dataset with sufficient examples of such facial hairstyles. Our interactive system allows for editing both the color and orientation of the hair, as well as providing additional tools to author varying styles such as sparse or dense hair. Despite the significant research in the domain of image editing, few prior works investigate high quality and intuitive synthesis of facial hair.
Though Brock~\etal~\cite{brock2016neural} allows for adding or editing the overall appearance of the subject's facial hair, their results lack details and can only operate on low-resolution images.
To the best of our knowledge, we present the first interactive framework that is specially tailored for synthesizing high-fidelity facial hair with large variations.

%% file: uist_submission/overview.tex
In Sec.~\ref{sec:method} we describe our network pipeline (Fig.~\ref{fig:pipeline}), the architectures of our networks, and the training process. Sec.~\ref{sec:dataset} describes the datasets we use, including the large synthetic dataset we generate for the initial stage of our training process, our dataset of real facial hair images we use for the final stage of training for refinement, and our method for annotating these images with the guide strokes used as input during training (Fig.~\ref{fig:dataset}). Sec.~\ref{sec:ui} describes the user interface tools we provide to allow for intuitive and efficient authoring of input data describing the desired hairstyle (Fig.~\ref{fig:ui}). Finally, Sec.~\ref{sec:results} provides sample results (Fig.~\ref{fig:results}), comparisons with alternative approaches (Figs.~\ref{fig:copy_paste} and~\ref{fig:comparison}), an ablation analysis of our architecture and training process (Table~\ref{tab:ablation}), and descriptions of the perceptual and user study we use to evaluate the quality of our results and the utility of our interface.

%% file: uist_submission/network.tex
\label{sec:method}

Given an image with a segmented region defining the area in which synthesis is to be performed, and a set of guide strokes, we use a two-stage inference process that populates the selected region of the input image with desired hairstyle as shown in Fig.~\ref{fig:pipeline}. The first stage synthesizes an initial approximation of the content of the segmented region, while the second stage refines this initial result and adjusts it to allow for appropriate compositing into the final image.
\vspace{-1mm}
\paragraph{Initial Facial Hair Synthesis.}
The input to the first network consists of a 1-channel segmentation map of the target region, and a 4-channel (RGBA) image of the provided guide strokes within this region. The output is a synthesized approximation of the hair in the segmented region. 

The generator network is an encoder-decoder architecture extending upon the image-to-image translation network of~\cite{isola2016image}. We extend the decoder architecture with a final 3x3 convolution layer, with a step size of 1 and 1-pixel padding, to refine the final output and reduce noise. To exploit the rough spatial correspondence between the guide strokes drawn on the segmented region of the target image and the expected output, we utilize skip connections~\cite{DBLP:journals/corr/RonnebergerFB15} to capture low-level details in the synthesized image.

We train this network using the $L_1$ loss between the ground-truth hair region and the synthesized output. We compute this loss only in the segmented region encouraging the network to focus its capacity on synthesizing the facial hair with no additional compositing constraints.
We also employ an adversarial loss \cite{goodfellow2014generative} by using a discriminator based on the architecture of~\cite{isola2016image}.  We use a conditional discriminator, which accepts both the input image channels and the corresponding synthesized or real image. This discriminator is trained in conjunction with the generator to determine whether a given hair image is real or synthesized, and whether it plausibly corresponds to the specified input. Finally, we use a perceptual loss metric~\cite{DBLP:journals/corr/JohnsonAL16,gatys2016image}, represented using a set of higher-level feature maps extracted from a pre-existing image classification network (\ie, VGG-19~\cite{Simonyan14c}). This is effective in encouraging the network to synthesize results with content that corresponds well with plausible images of real hair. The final loss $L(I_s, I_{gt})$ between the synthesized ($I_s$) and ground truth facial hair images ($I_{gt}$) is thus:
\begin{equation}
    L_{f} (I_s, I_{gt}) = \omega_1 L_1(I_s, I_{gt}) + \omega_{adv} L_{adv}(I_s, I_{gt}) + \omega_{per}L_{per}(I_s, I_{gt}),
\end{equation}
where $L_1$, $L_{adv}$, and $L_{per}$ denote the $L_1$, adversarial, and perceptual losses respectively. The relative weighting of these losses is determined by $\omega_1$, $\omega_{adv}$, and $\omega_{per}$. We set these weights ($\omega_1=50$, $\omega_{adv}=1$, $\omega_{per}=0.1$), such that the average gradient of each loss is at the same scale. We first train this network until convergence on the test set using our large synthetic dataset (see Sec.~\ref{sec:dataset}). It is then trained in conjunction with the refinement/compositing network on the smaller real image dataset to allow for better generalization to unconstrained real-world images.

\vspace{-1mm}
\paragraph{Refinement and Compositing.}
Once the initial facial hair region is synthesized, we perform refinement and compositing into the input image. This is achieved by a second encoder-decoder network. The input to this network is the output of the initial synthesis stage, the corresponding segmentation map, and the segmented target image (the target image with the region to be synthesized covered by the segmentation mask). The output is the image with the synthesized facial hair refined and composited into it.

The architecture of the second generator and discriminator networks are identical to the first network, with only the input channel sizes adjusted accordingly. While we use the adversarial and perceptual losses in the same manner as the previous stage, we define the $L_1$ loss on the entire synthesized image. However, we increase the weight of this loss by a factor of 0.5 in the segmented region containing the facial hair. The boundary between the synthesized facial hair region and the rest of the image is particularly important for plausible compositions. Using erosion/dilation operations on the segmented region (with a kernel size of 10 for each operation), we compute a mask covering this boundary. We further increase the weight of the loss for these boundary region pixels by a factor of 0.5. More details on the training process can be found in the appendix.

%% file: uist_submission/dataset.tex
To train a network to synthesize realistic facial hair, we need a sufficient number of training images to represent the wide variety of existing facial hairstyles (\eg, varying shape, length, density, and material and color properties), captured under varying conditions (\eg, different viewpoints). We also need a method to represent the distinguishing features of these hairstyles in a simple and abstract manner that can be easily replicated by a novice user.
\vspace{-3mm}

\begin{figure}[ht!]
\centering
    \includegraphics[width=0.85\columnwidth]{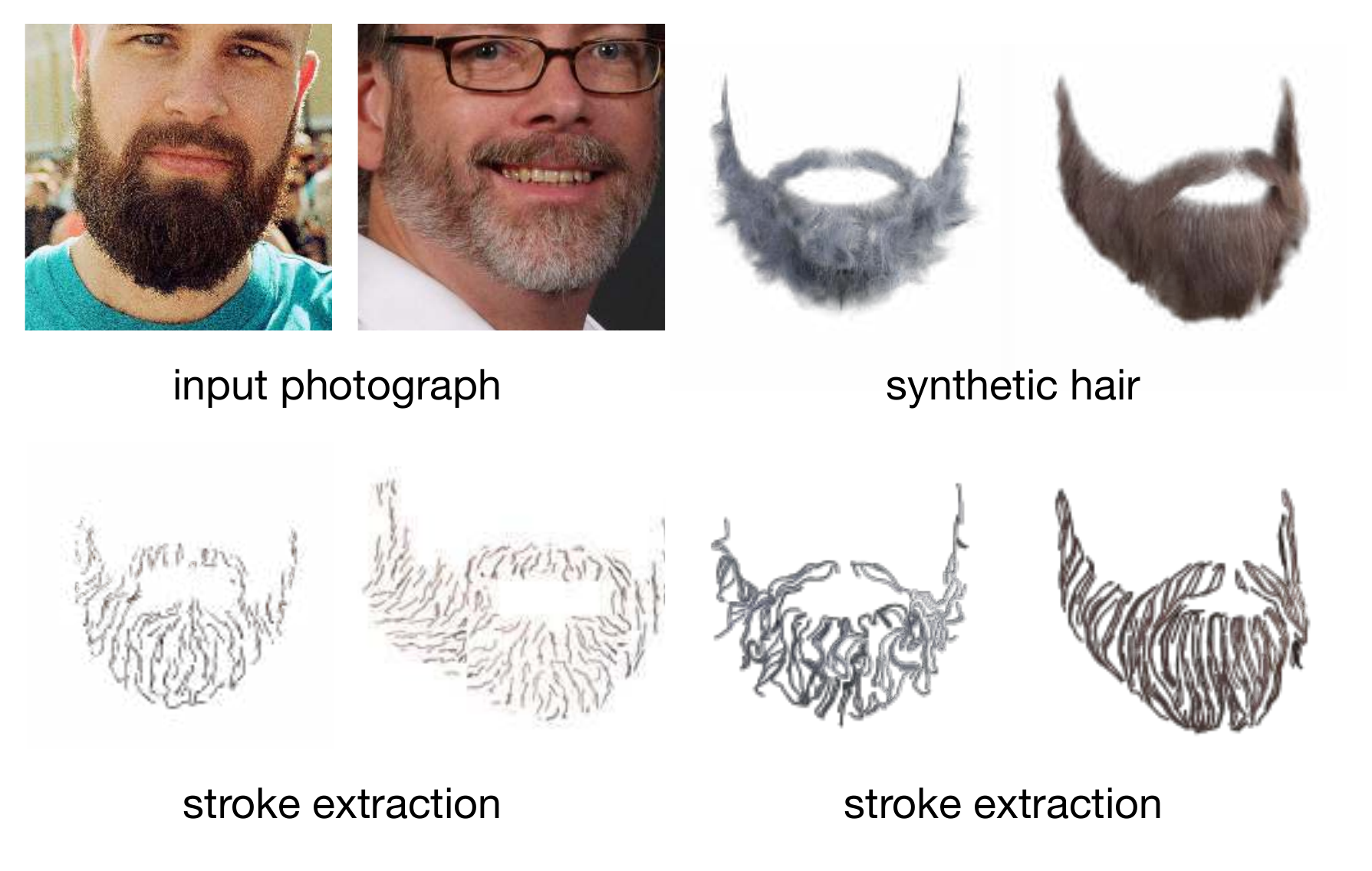}
  \vspace{-4mm}
  \caption{We train our network with both real (row 1, columns 1-2) and synthetic (row 1, columns 3-4) data. For each input image we have a segmentation mask denoting the facial hair region, and a set of guide strokes (row 2) that define the hair's local structure and appearance.}
  \label{fig:dataset}
  \vspace{-3mm}
\end{figure}

\vspace{-4mm}

\paragraph{Data collection}
To capture variations across different facial hairstyles in a controlled manner, we generate a large-scale synthetic dataset using the Whiskers plugin \cite{daz3dwhiskers} provided for the Daz 3D modeling framework \cite{daz3d}. This plugin provides $50$ different facial hairstyles (\eg full beards, moustaches, goatees), with parameters controlling the color and length of the selected hairstyle. The scripting interface provided by this modeling framework allows for programmatically adjusting the aforementioned parameters and rendering the corresponding images. By rendering the alpha mask for the depicted facial hair, we automatically extract the corresponding segmentation map. For each facial hairstyle, we synthesize it at $4$ different lengths and $8$ different colors. We render each hairstyle from $19$ viewpoints sampled by rotating the 3D facial hair model around its central vertical axis in the range $[-90\degree, 90\degree]$ at $10\degree$ intervals, where $0\degree$ corresponds to a completely frontal view and $90\degree$ corresponds to a profile view (see Fig.~\ref{fig:dataset}, columns 3-4 for examples of these styles and viewpoints). We use the Iray~\cite{DBLP:journals/corr/KellerWRSAKK17} physically-based renderer to generate $30400$ facial hair images with corresponding segmentation maps. 

To ensure our trained model generalizes to real images, we collect and manually segment the facial hair region in a small dataset of such images (approximately 1300 images) from online image repositories containing a variety of styles, \eg short, stubble, long, dense, curly, and straight, and large variations in illumination, pose and skin color.
\vspace{-3mm}

\paragraph{Dataset Annotation}
Given input images with masks denoting the target region to be synthesized, we require guide strokes providing an abstract representation of the desired facial hair properties (\eg, the local color and shape of the hair).
We simulate guide strokes by integrating a vector field computed based on the approach of ~\cite{Kyprianidis:2011:CEF}, which computes the dominant local orientation from the per-pixel structure tensor, then produces abstract representations of images by smoothing them using line integral convolution in the direction of minimum change.
Integrating at points randomly sampled in the vector field extracted from the segmented hair region in the image produces guide lines that resemble the types of strokes specified by the users. These lines generally follow prominent wisps or strands of facial hair in the image (see Fig.~\ref{fig:dataset}).

%% file: uist_submission/ui_cvpr.tex
We provide an interactive user interface with tools to perform intuitive facial hair editing and synthesis in an arbitrary input image. The user first specifies the hair region via the mask brush in the input image,
then draws guide strokes within the mask abstractly describing the overall desired hairstyle. Our system provides real-time synthesized resulted after each edit to allow for iterative refinement with instant feedback.
Please refer to the supplementary video for example sessions and the appendix for more details on our user interface.
Our use of guide strokes extracted from vector fields during training enables the use of various intuitive and lightweight tools to facilitate the authoring process. In addition, the generative power of our network allows for synthesizing a rough initial approximation of the desired hairstyle with minimal user input.
\vspace{-3mm}

\begin{figure}[h!]
\centering
  \includegraphics[width=\columnwidth]{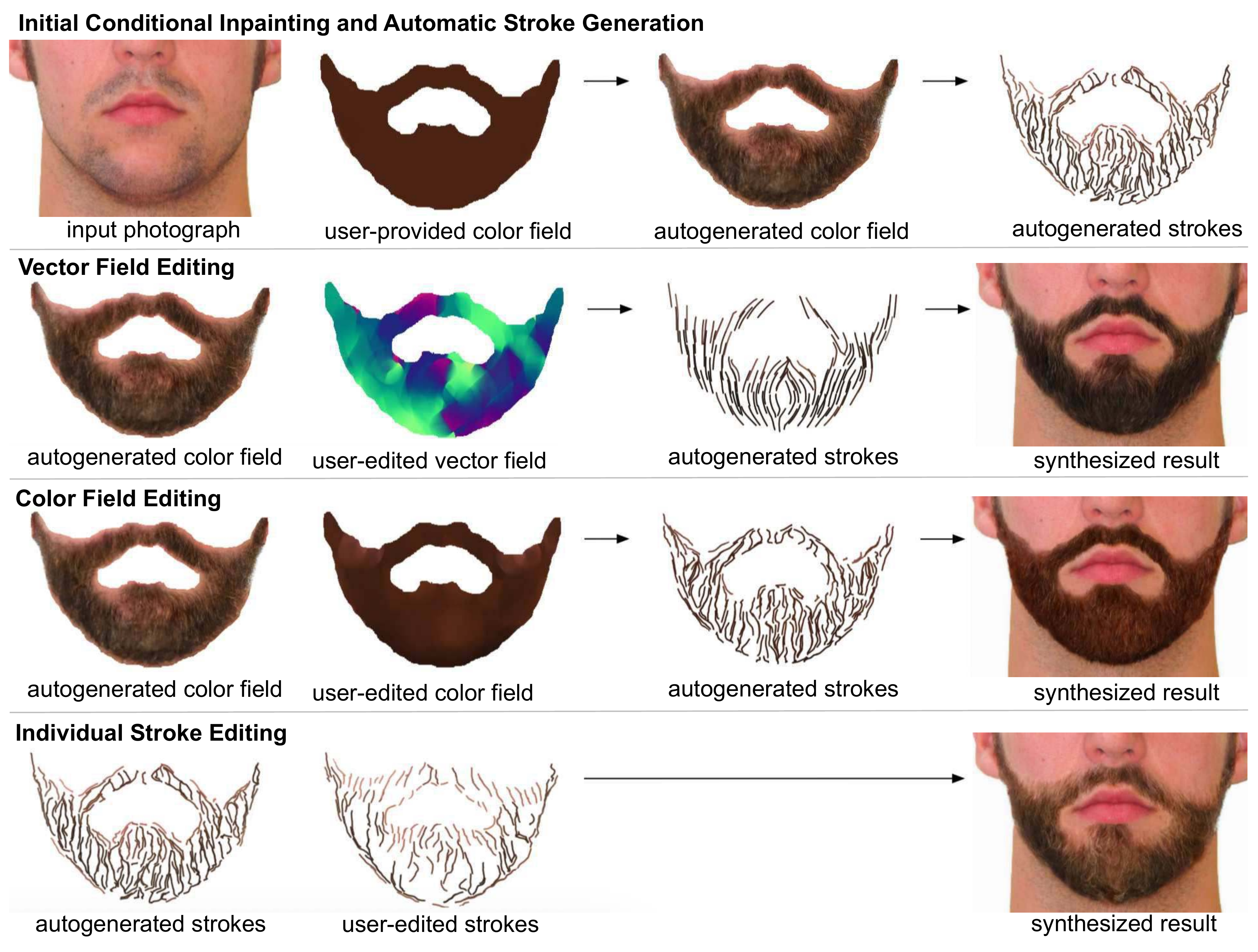}
  \vspace{-4mm}
  \caption{Editing examples. Row 1: Synthesizing an initial estimate given a user-specified mask and color. Extracting the vector field from the result allows for creating an initial set of strokes that can be then used to perform local edits. Row 2: Editing the extracted vector field to change the facial hair structure while retaining the overall color. Row 3: Changing the color field while using the vector field from the initial synthesis result allows for the creation of strokes with different colors but similar shapes to the initially generated results. Row 4: Editing the strokes extracted from the initial synthesis result allows for subtle updates, \eg making the beard sparser around the upper cheeks.}
  \label{fig:ui}
  \vspace{-3mm}
\end{figure}

\paragraph{Guide stroke initialization.}

We provide an optional initialization stage where an approximation of the desired hairstyle is generated given only the input image, segmentation mask, and a corresponding color for the masked region. This is done by adapting our training procedure to train a separate set of networks with the same architectures described in Sec.~\ref{sec:method} using this data without the aforementioned guide strokes.
Given a segmented region and the mean RGB color in this region, the network learns a form of conditional inpainting, synthesizing appropriate facial hair based on the region's size, shape, color, and the context provided by the unmasked region of the image. For example, using small masked regions with colors close to the surrounding skin tone produces sparse, short facial hair, while large regions with a color radically different from the skin tone produces longer, denser hairstyles.
The resulting facial hair is realistic enough to extract an initial set of guide strokes from the generated image as is done with real images (see Sec.~\ref{sec:dataset}).
Fig.~\ref{fig:ui} (top row) demonstrates this process.
\vspace{-3mm}

\paragraph{Guide stroke editing}

These initial strokes provide a reasonable initialization the user's editing.
The vector field used to compute these strokes and the initial synthesis result, which acts as the underlying color field used to compute the guide stroke color, can also be edited. As they are changed, we automatically repopulate the edited region with strokes corresponding to the specified changes.
We provide brush tools to make such modifications to the color or vector fields, as seen in Fig.~\ref{fig:ui}.
The user can adjust the brush radius to alter the size of the region affected region, as well as the intensity used when blending with previous brush strokes. 
The users can also add, delete, or edit the color of guide strokes to achieve the desired alterations.

%% file: uist_submission/results.tex
\begin{figure*}[t!]
  \centering
  \includegraphics[width=1.0\textwidth]{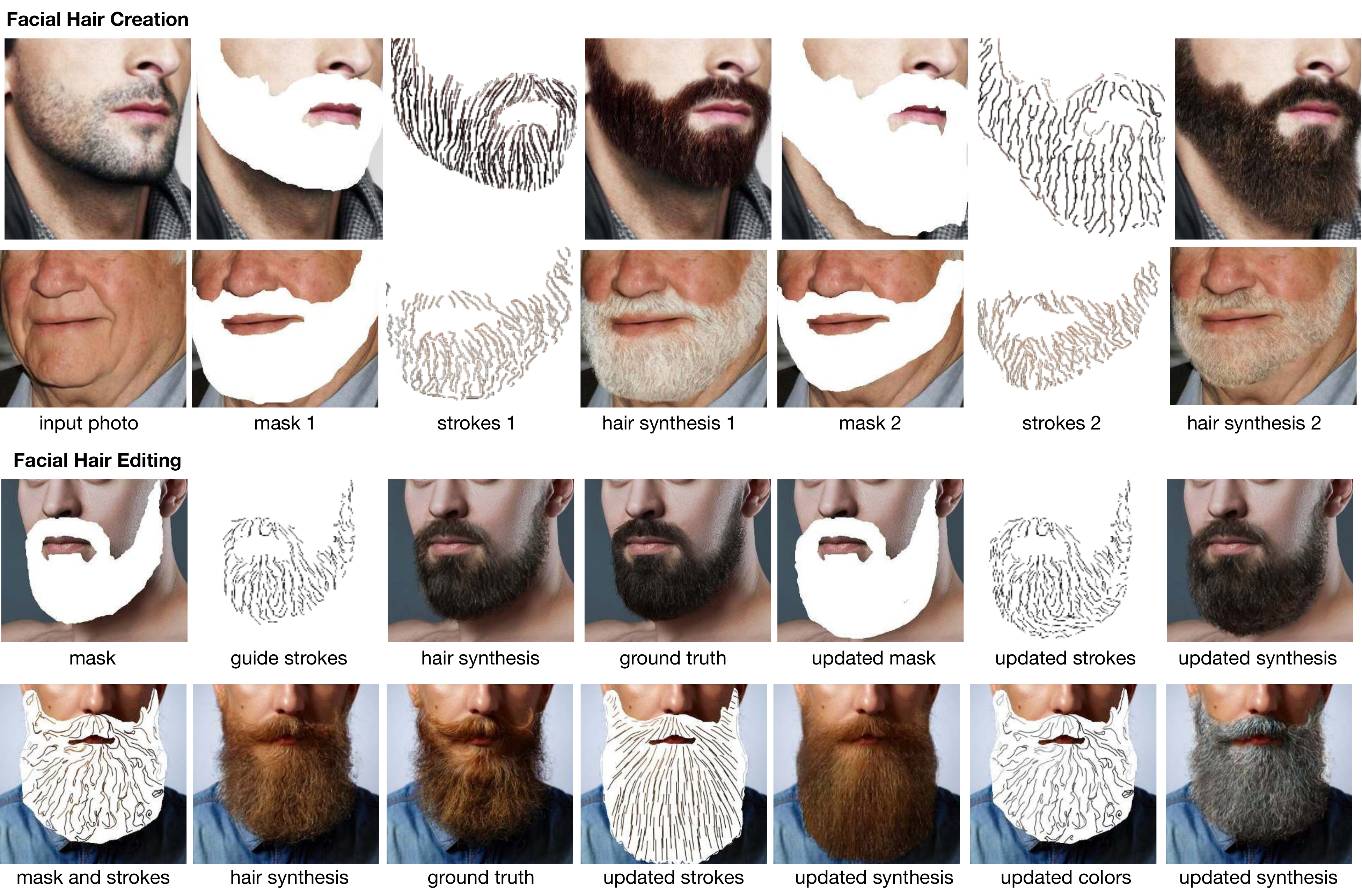}
  \vspace{-7mm}
  \caption{Example results. We show several example applications, including creating and editing facial hair on subjects with no facial hair, as well as making modifications to the overall style and color of facial hair on bearded individuals.}
    \label{fig:results}
    \vspace{-1mm}
\end{figure*}

We show various examples generated by our system in Figs.~\ref{fig:teaser} and~\ref{fig:results}. It can be used to synthesize hair from scratch (Fig.~\ref{fig:results}, rows 1-2) or to edit existing facial hair (Fig.~\ref{fig:results}, rows 3-4). As shown, our system can generate facial hair of varying overall color (red vs. brown), length (trimmed vs. long), density (sparse vs. dense), and style (curly vs. straight). Row 2 depicts a complex example of a white, sparse beard on an elderly subject, created using light strokes with varying transparency. By varying these strokes and the masked region, we can generate a relatively long, mostly opaque style (column 4) or a shorter, stubbly and more translucent style (column 7).
Please consult the supplementary video for live recordings of several editing sessions and timing statistics for the creation of these example results.

\vspace{-3mm}
\paragraph{Perceptual study}
To evaluate the perceived realism of the editing results generated by our system, we conducted a perceptual study in which 11 subjects viewed 10 images of faces with only real facial hair and 10 images with facial hair manually created using our method, seen in a random order. Users observed each image for up to 10 seconds and decided whether the facial hair was real or fake/synthesized. Real images were deemed real 80\% of the time, while edited images were deemed real 56\% of the time. In general, facial hair synthesized with more variation in color, texture, and density were perceived as real, demonstrating the benefits of the local control tools in our interface. Overall, our system's results were perceived as generally plausible by all of the subjects, demonstrating the effectiveness of our method.

\vspace{-3mm}
\paragraph{Comparison with naive copy-paste}
\begin{figure}[h]
 \includegraphics[width=\columnwidth]{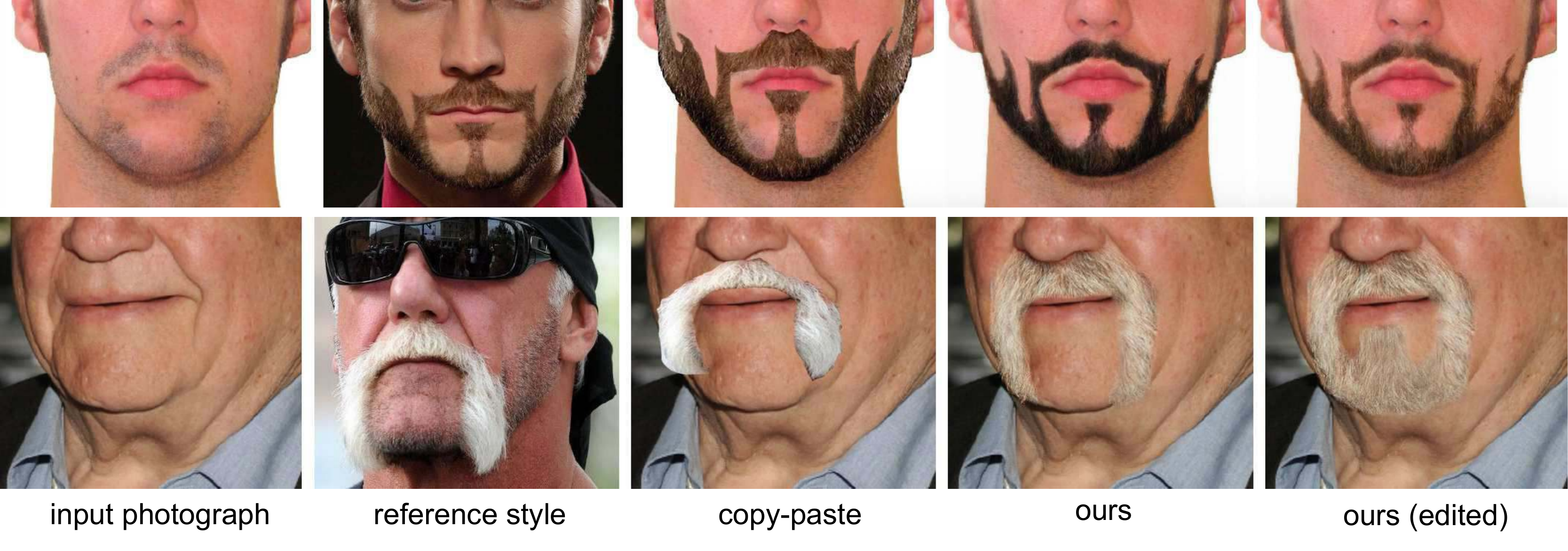}
  \vspace{-6mm}
  \caption{Comparison to naive copy-pasting images from reference photographs. Aside from producing more plausible results, our approach enables editing the hair's color (row 1, column 5) and shape (row 2, column 5).}
  \label{fig:copy_paste}
  \vspace{1mm}
\end{figure}

A straightforward solution to facial hair editing is to simply copy similar facial hairstyles from a reference image. While this may work for reference images depicting simple styles captured under nearly identical poses and lighting conditions to those in the the target photograph, slight disparities in these conditions result in jarring incoherence between the copied region and the underlying image. In contrast, our method allows for flexible and plausible synthesis of various styles, and enables the easy alteration of details, such as the shape and color of the style depicted in the reference photograph to allow for more variety in the final result. See Fig.~\ref{fig:copy_paste} for some examples of copy-pasting vs. our method. Note that when copy-pasting, the total time to cut, paste, and transform (rotate and scale) the copied region to match the underlying image was in the range of 2-3 minutes, which is comparable to the amount of time spent when using our method. 

\vspace{-3mm}
\paragraph{Comparison with texture synthesis}

\begin{figure}[h!]
  \centering
  \vspace{1mm}
  \includegraphics[width=\columnwidth]{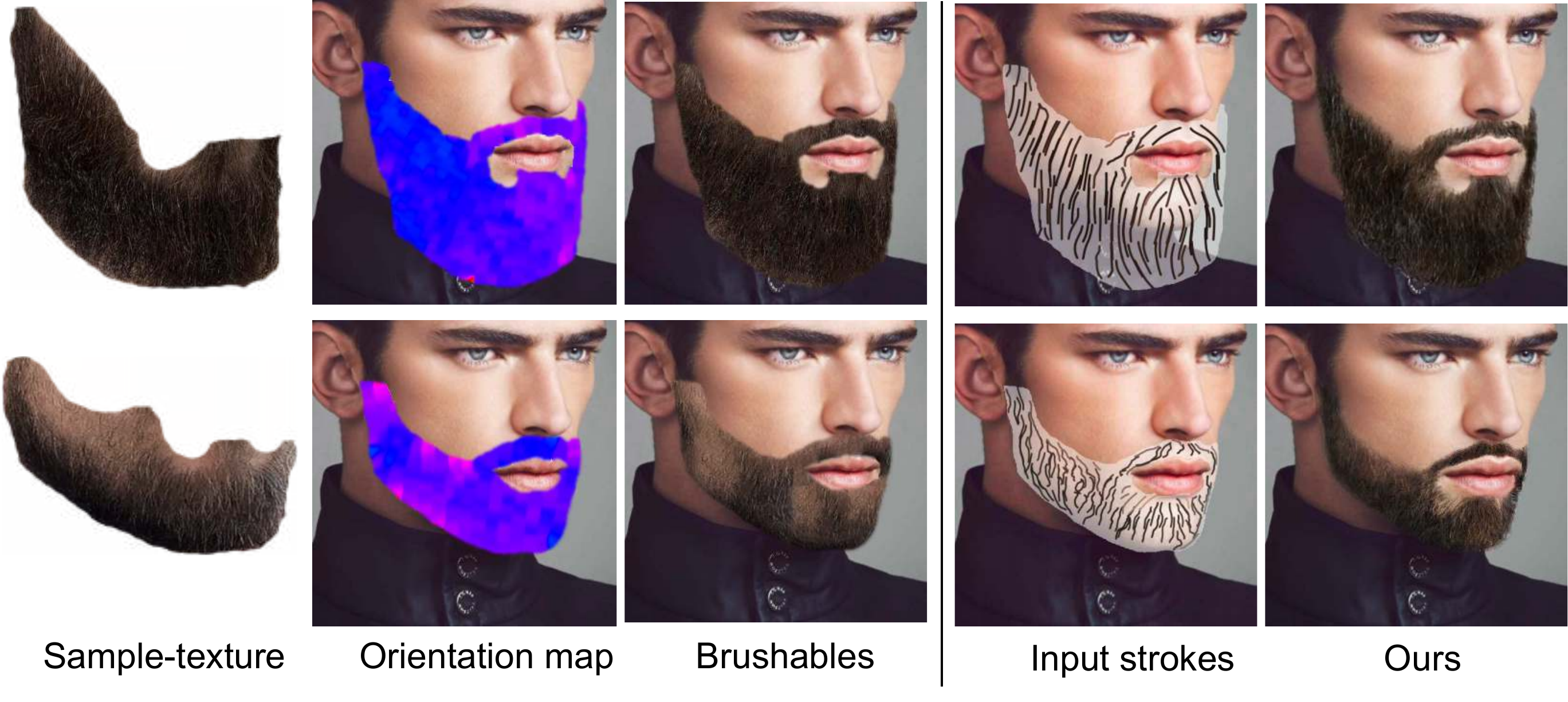}
  \vspace{-7mm}
  \caption{Results of our comparison with Brushables.}
  \vspace{-3mm}
{}
  \label{fig:comparison}
\end{figure}
 
We compare our method to Brushables~\cite{Lukac:2015:BEE}, which has an intuitive interface for orientation maps to synthesize images that match the target shape and orientation while maintaining the textural details of a reference image. We can use Brushables to synthesize facial hair by providing it with samples of a real facial hair image and orientation map, as shown in Fig.~\ref{fig:comparison}. For comparison, with our system we draw strokes in the same masked region on the face image. While Brushables synthesizes hair regions matching the provided orientation map, our results produce a more realistic hair distribution and appear appropriately volumetric in nature. Our method also handles skin tones noticeably better near hair boundaries and sparse, stubbly regions. Our method takes 1-2 seconds to process each input operation, while the optimization in Brushables takes 30-50 seconds for the same image size.
\vspace{-1mm}

\paragraph{Ablation analysis}

\begin{table}
\small
\centering
\setlength{\tabcolsep}{2pt}
\begin{tabular}{l c c c c c c}
\toprule
& L1$\downarrow$ & VGG$\downarrow$ & MSE$\downarrow$ & PSNR$\uparrow$ & SSIM$\uparrow$ & FID$\downarrow$ \\
Isola~\etal~\cite{isola2016image} & 0.0304 & 168.5030 & 332.69 & 23.78 & 0.66 & 121.18 \\
Single Network & 0.0298 & 181.75 & \textbf{274.88} & 24.63 & 0.67 & 75.32 \\
Ours, w/o GAN & 0.0295 & 225.75 & 334.51 & \textbf{24.78} & \textbf{0.70} & 116.42 \\
Ours, w/o VGG & 0.0323 & 168.3303 & 370.19 & 23.20 & 0.63 & 67.82 \\
Ours, w/o synth. & 0.0327 & 234.5 & 413.09 & 23.55 & 0.62 & 91.99 \\
Ours, only synth. & 0.0547 & 235.6273 & 1747.00 & 16.11 & 0.60 & 278.17 \\
Ours & \textbf{0.0275} & \textbf{119.00} & 291.83 & 24.31 & 0.68 & \textbf{53.15} \\
\bottomrule
\end{tabular}
\vspace{-3mm}
\caption{Quantitative ablation analysis.
\vspace{-6mm}
}
\label{tab:ablation}
\end{table}
\begin{figure*}[h!]
    \centering
        \subfloat[Input]{\begin{tabular}[b]{c}\includegraphics[width=0.100\linewidth]{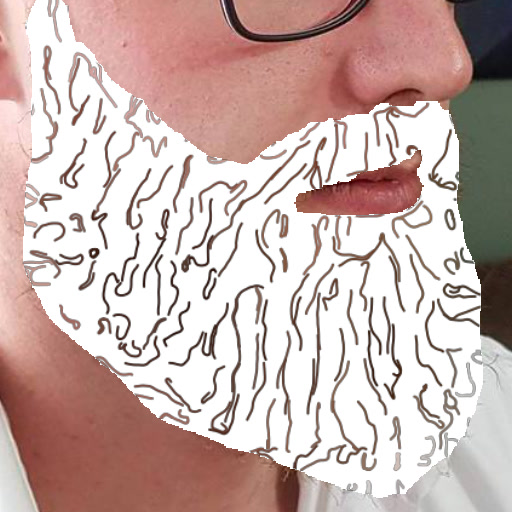}\\        
\includegraphics[width=0.100\linewidth]{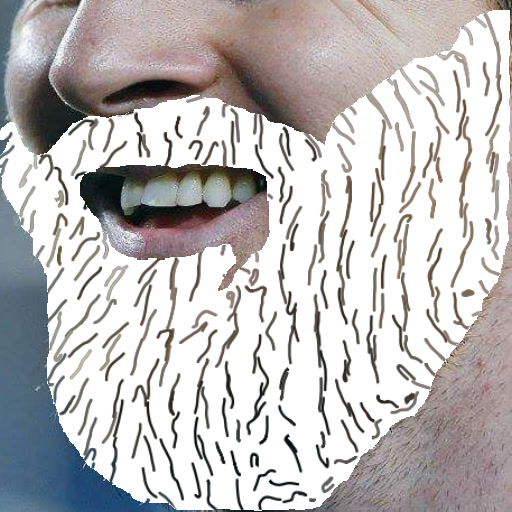}
\end{tabular}}\hspace{-10pt}
        \subfloat[Isola~\etal]{\begin{tabular}[b]{c}\includegraphics[width=0.100\linewidth]{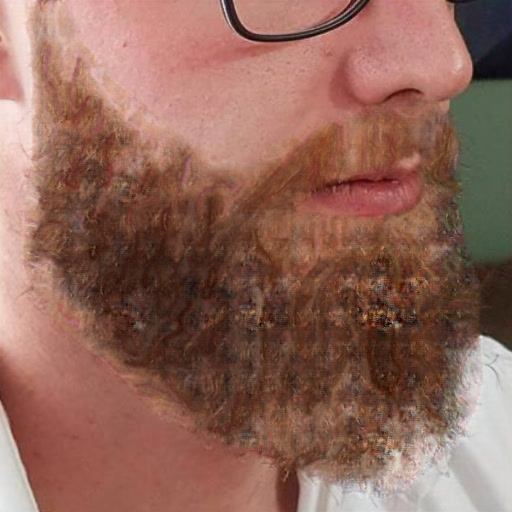}\\        
\includegraphics[width=0.100\linewidth]{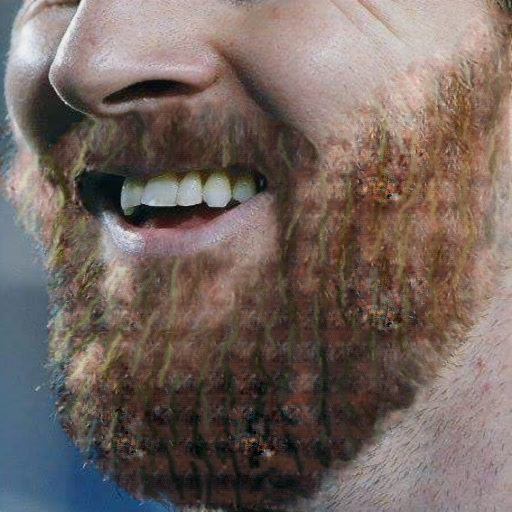}
\end{tabular}}\hspace{-10pt}
        \subfloat[Single Network]{\begin{tabular}[b]{c}\includegraphics[width=0.100\linewidth]{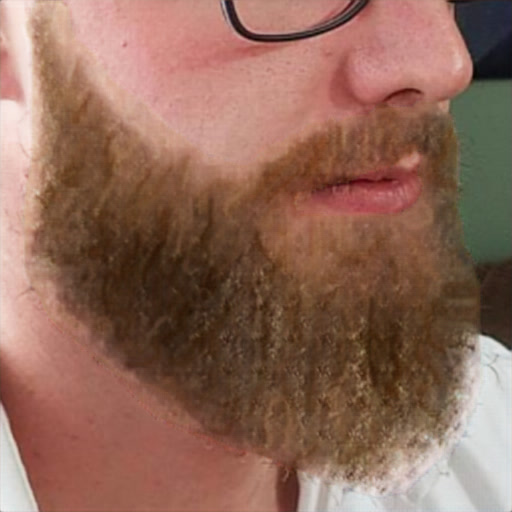}\\        
\includegraphics[width=0.100\linewidth]{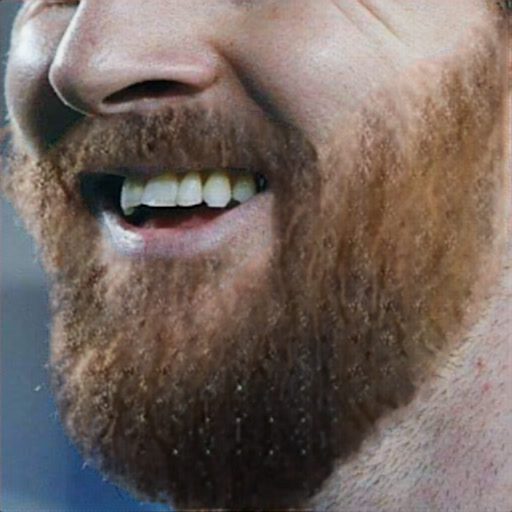}
\end{tabular}}\hspace{-10pt}
        \subfloat[w/o GAN]{\begin{tabular}[b]{c}\includegraphics[width=0.100\linewidth]{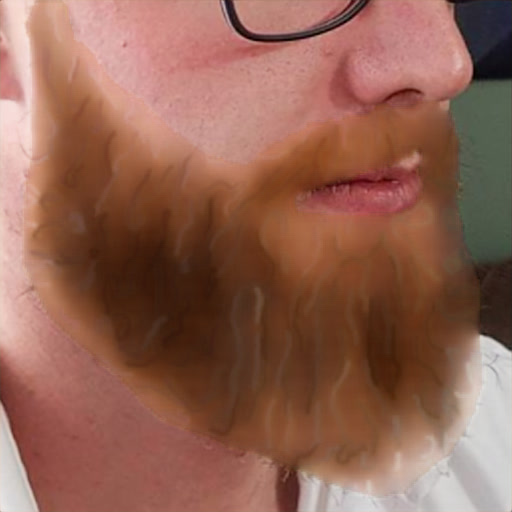}\\        
\includegraphics[width=0.100\linewidth]{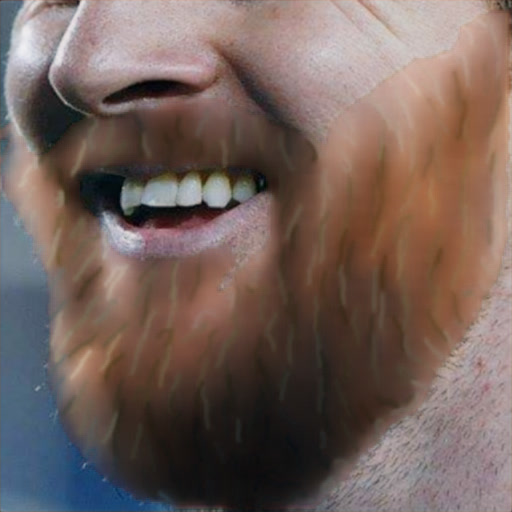}
\end{tabular}}\hspace{-10pt}
        \subfloat[w/o VGG]{\begin{tabular}[b]{c}\includegraphics[width=0.100\linewidth]{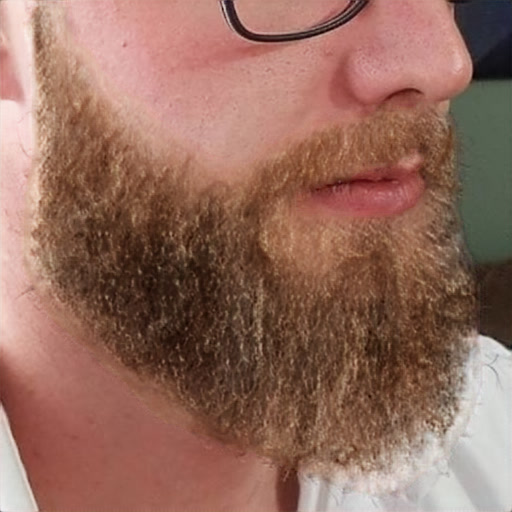}\\        
\includegraphics[width=0.100\linewidth]{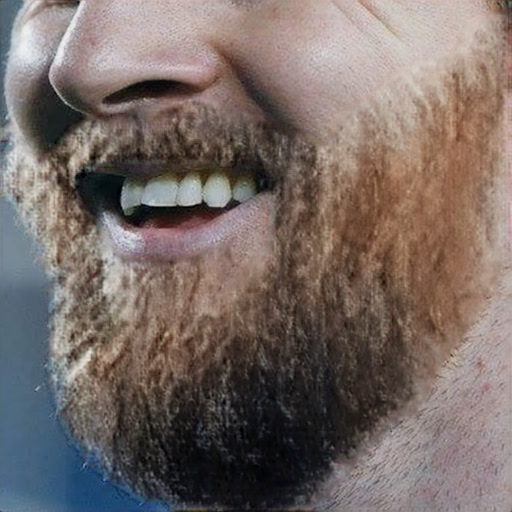}
\end{tabular}}\hspace{-10pt}
        \subfloat[w/o synth. data]{\begin{tabular}[b]{c}\includegraphics[width=0.100\linewidth]{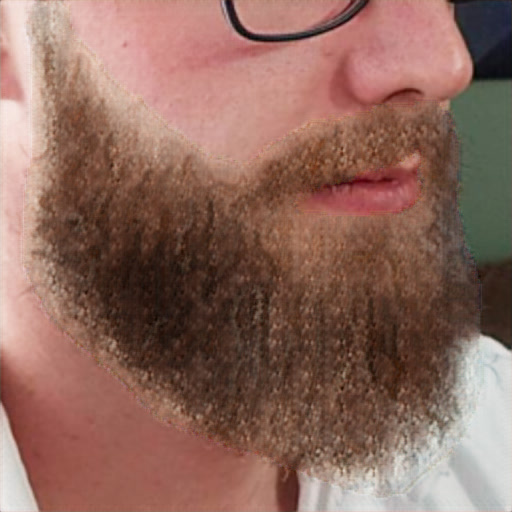}\\        
\includegraphics[width=0.100\linewidth]{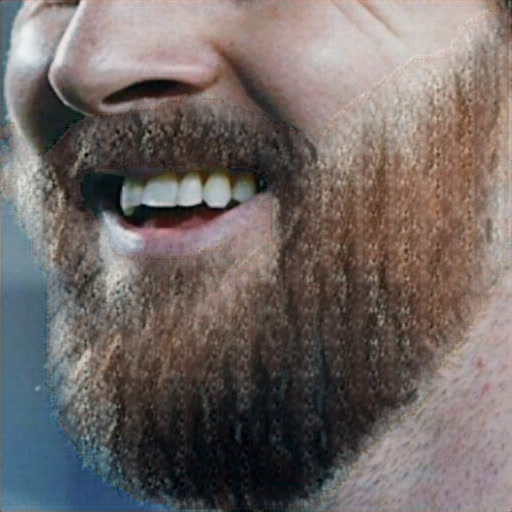}
\end{tabular}}\hspace{-10pt}
        \subfloat[Ours, final]{\begin{tabular}[b]{c}\includegraphics[width=0.100\linewidth]{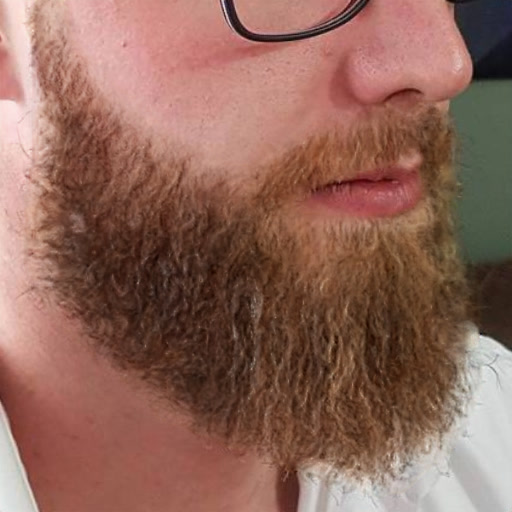}\\        
\includegraphics[width=0.100\linewidth]{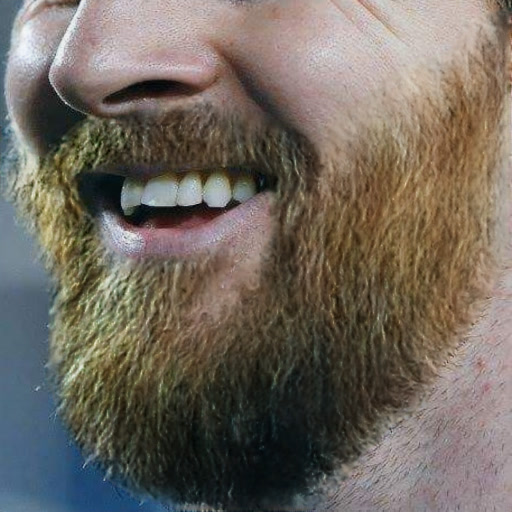}
\end{tabular}}\hspace{-10pt}
        \subfloat[Ground truth]{\begin{tabular}[b]{c}\includegraphics[width=0.100\linewidth]{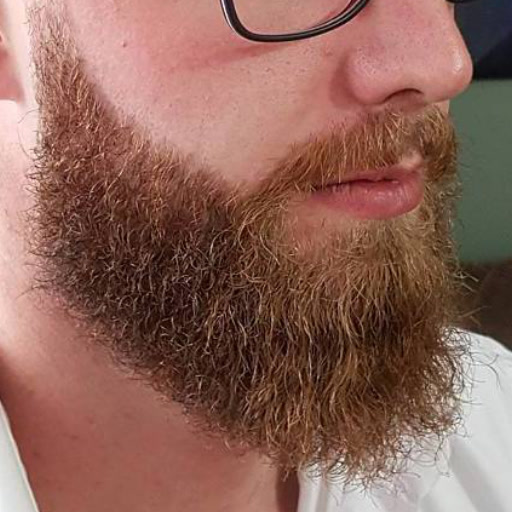}\\     
\includegraphics[width=0.100\linewidth]{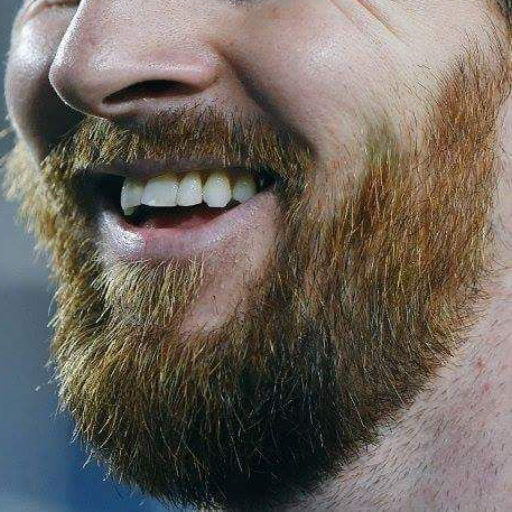}
\end{tabular}} \vspace{-2mm}
        \caption{Qualitative comparisons for ablation study.}
  {}
  \label{fig:ablation}
  \vspace{-3mm}
\end{figure*}

\begin{figure}[h!]
  \centering
  \includegraphics[width=\columnwidth]{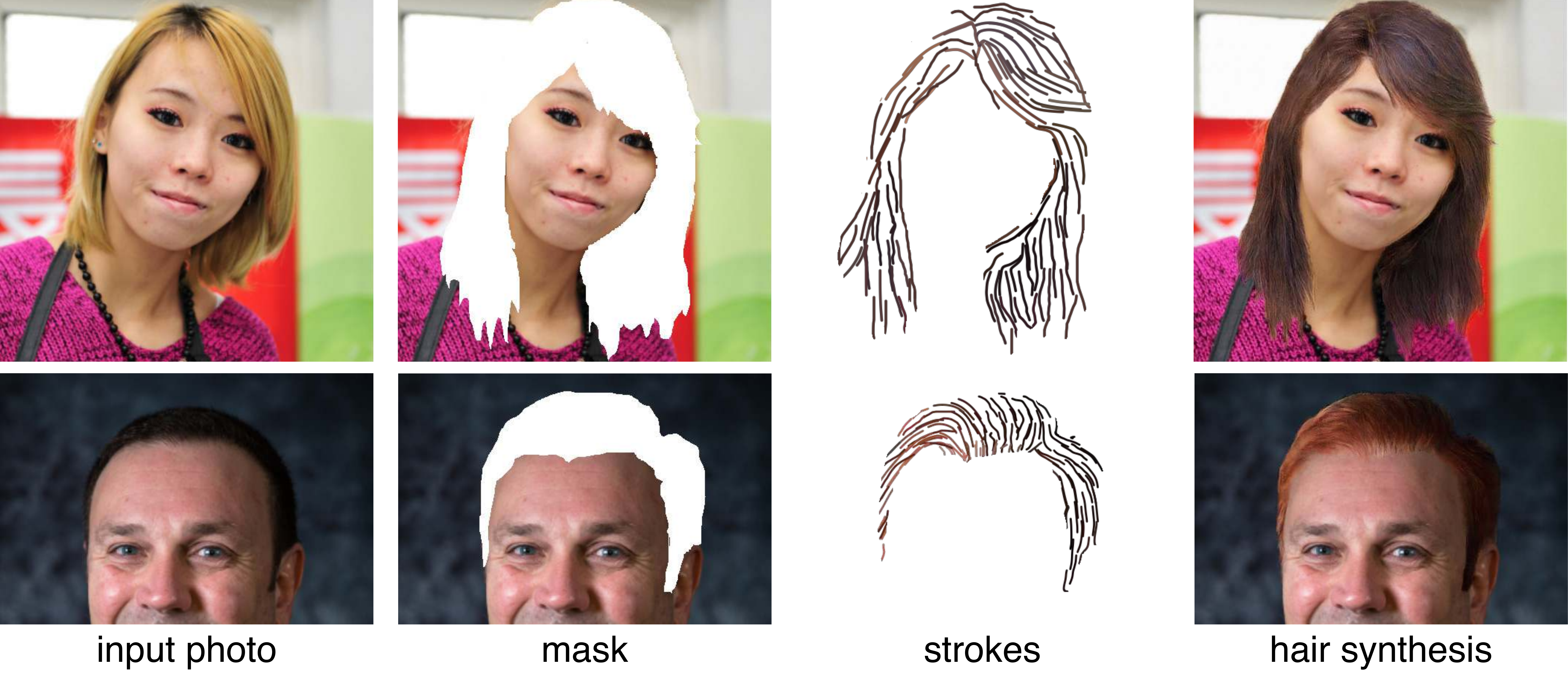}
  \vspace{-7mm}
  \caption{Scalp hair synthesis and compositing examples.}
    \label{fig:scalp_results}
    \vspace{-4mm}
\end{figure}
As described in Sec.~\ref{sec:method}, we use a two-stage network pipeline trained with perceptual and adversarial loss, trained with both synthetic and real images. We show the importance of each component with an ablation study.
With each component, our networks produce much higher quality results than the baseline network of \cite{isola2016image}.

A quantitative comparison is shown in Table~\ref{tab:ablation}, in which we summarize the loss values computed over our test data set using several metrics, using 100 challenging ground truth validation images not used when computing the training or testing loss. While using some naive metrics variations on our approach perform comparably well to our final approach, we note that ours outperforms all of the others in terms of the Fr\'echet Inception Distance (FID)~\cite{DBLP:journals/corr/HeuselRUNKH17}, as well as the MSE loss on the VGG features computed for the synthesized and ground truth images. This indicates that our images are perceptually closer to the actual ground truth images. Selected qualitative examples of the results of this ablation analysis can be seen in Fig.~\ref{fig:ablation}. More can be found in the appendix.
\vspace{-3mm}

\paragraph{User study.}
We conducted a preliminary user study to evaluate the usability of our system. The study included 8 users, of which one user was a professional technical artist.
The participants were given a reference portrait image and asked to create similar hair on a different clean-shaven subject via our interface. Overall, participants were able to achieve reasonable results.
From the feedback, the participants found our system novel and useful. When asked what features they found most useful, some users commented that they liked the ability to create a rough approximation of the target hairstyle given only a mask and average color. Others strongly appreciated the color and vector field brushes, as these allowed them to separately change the color and structure of the initial estimate, and to change large regions of the image without drawing each individual stroke with the appropriate shape and color. 
Please refer to the appendix for the detailed results of the user study and example results created by the participants.
\vspace{-3mm}

\paragraph{Application to non-facial hair.}

While we primarily focus on the unique challenges of synthesizing and editing facial hair in this work, our method can easily be extended to scalp hair with suitable training data. To this end, we refine our networks trained on facial hair with an additional training stage using 5320 real images with corresponding scalp hair segmentations, much in the same manner as we refine our initial network trained on synthetic data. This dataset was sufficient to obtain reasonable scalp synthesis and editing results.
See Fig.~\ref{fig:scalp_results} for scalp hair generation results.
Interestingly, this still allows for the synthesis of plausible facial hair along with scalp hair within the same target image using the same trained model, given appropriately masks and guide strokes.
Please consult the appendix for examples and further details.

%% file: uist_submission/conclusion.tex
While we demonstrate impressive results, our approach has several limitations. As with other data-driven algorithms, our approach is limited by the amount of variation found in the training dataset. Close-up images of high-resolution complex structures fail to capture all the complexity of the hair structure, limiting the plausibility of the synthesized images. As our training datasets mostly consist of images of natural hair colors, using input with very unusual hair colors also causes noticeable artifacts.
See Fig.~\ref{fig:limitation} for examples of these limitations.

We demonstrate that our approach, though designed to address challenges specific to facial hair, synthesizes compelling results when applied to scalp hair given appropriate training data.
It would be interesting to explore how well this approach extends to other related domains such as animal fur, or even radically different domains such as editing and synthesizing images or videos containing fluids or other materials for which vector fields might serve as an appropriate abstract representation of the desired image content.

%% file: uist_submission/supplementary.tex
\section{Interactive Editing}
\label{sec:supp_ui}
\input{uist_submission/ui.tex}

\begin{figure*}[h!]
\includegraphics[width=\textwidth]{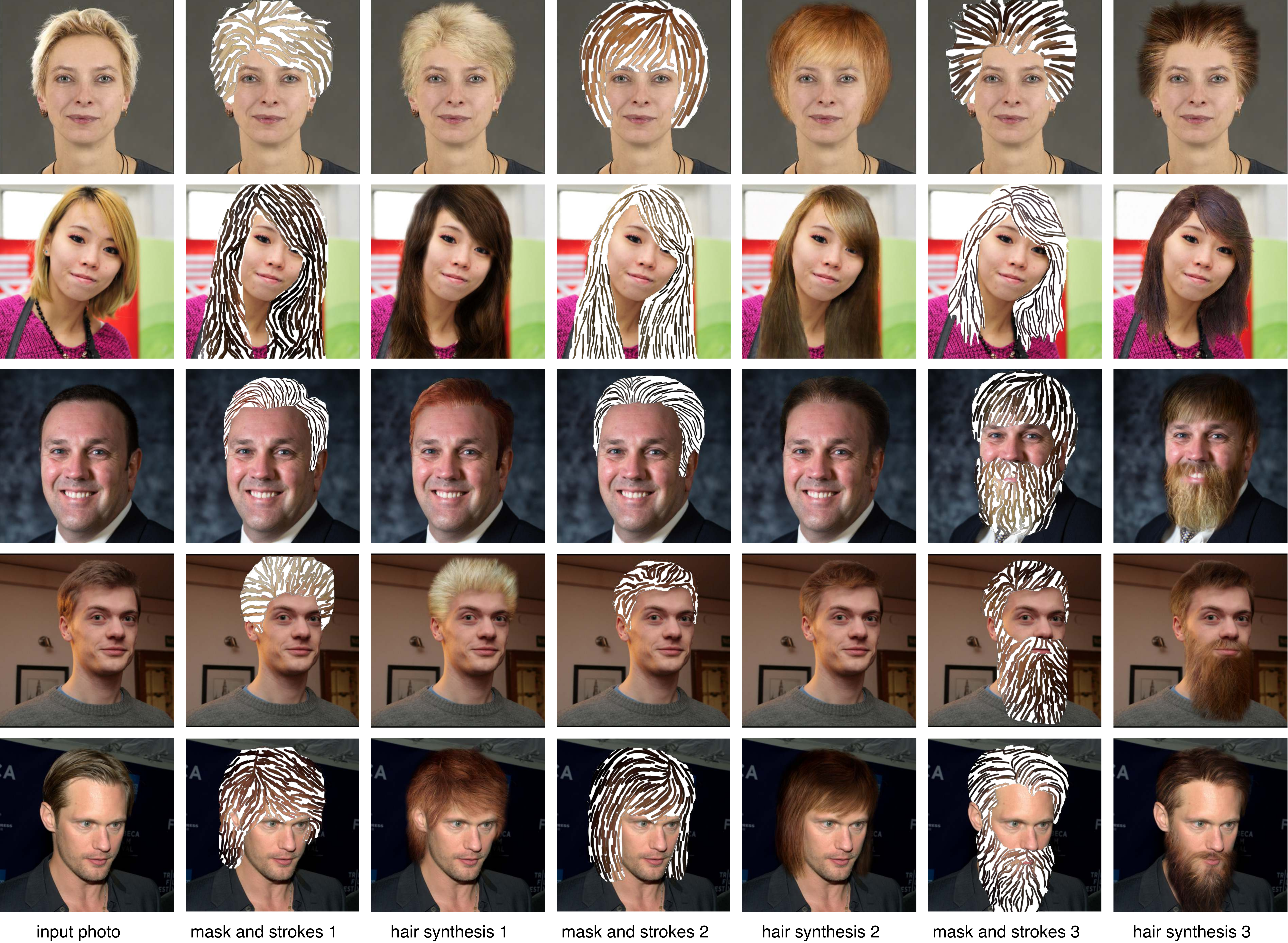}
\caption{Example results for synthesizing scalp hair, both with and without facial hair. Rows 1-2 shows examples for female subjects, while rows 3-5 depict male subjects. For the male subjects, columns 6-7 depict input and output to synthesize facial hair with scalp hair. These results are generated using the same models trained on a combination of facial and scalp hair.
\label{fig:supp_scalp_results}}
\end{figure*}

\section{Scalp and Facial Hair Synthesis Results}
\label{sec:scalp_results}
As described in Sec.~\ref{sec:results} and displayed in Fig.~\ref{fig:scalp_results}, we found that introducing segmented scalp hair with corresponding guide strokes (automatically extracted as described in Sec.~\ref{sec:dataset}) allows for synthesizing high-quality scalp hair and plausible facial hair with a single pair of trained networks to perform initial synthesis, followed by refinement and compositing.

We use 5320 real images with segmented scalp hair regions for these experiments. We do this by adding a second end-to-end training stage as described in Sec.~\ref{sec:implementation} in which the two-stage network pipeline is refined using only these scalp hair images. Interestingly, simply using the the real scalp and facial hair dataset simultaneously did not produce acceptable results. This suggests that the multi-stage refinement process we used to adapt our synthetic facial hair dataset to real facial hair images is also useful for further adapting the trained model to more general hairstyles.

Fig.~\ref{fig:supp_scalp_results} portrays several additional qualitative results from these experiments (all other results seen in the paper and supplementary video, with the exception of Fig.~\ref{fig:scalp_results}, were generated using a model trained using only the synthetic and real facial hair datasets). As can be seen, we can synthesize a large variety of hairstyles with varying structure and appearance for both female (rows 1-2) and male (rows 3-5) subjects using this model, and can synthesize both scalp and facial hair simultaneously (rows 3-5, columns 6-7). Though this increased flexibility in the types of hair that can be synthesized using this model comes with a small decrease in quality in some types of facial hair quite different from that seen in the scalp hair database (\eg, the short, sparse facial hair seen in Fig.~\ref{fig:results}, row 2, column 7, which was synthesized using a model trained using only facial hair images), relatively dense facial hairstyles such as those portrayed in Fig.~\ref{fig:supp_scalp_results} can still be plausibly synthesized simultaneously with a wide variety of scalp hairstyles.

\section{Ablation Study Results}
\label{sec:ablation_results}
\begin{figure*}[h!]
    \centering
        \subfloat[Input]{\begin{tabular}[b]{c}\includegraphics[width=0.1125\linewidth]{figs/raster/ablate/input/multistroke_stream_00_000012_img_01_composite.jpg}\\        
            \includegraphics[width=0.1125\linewidth]{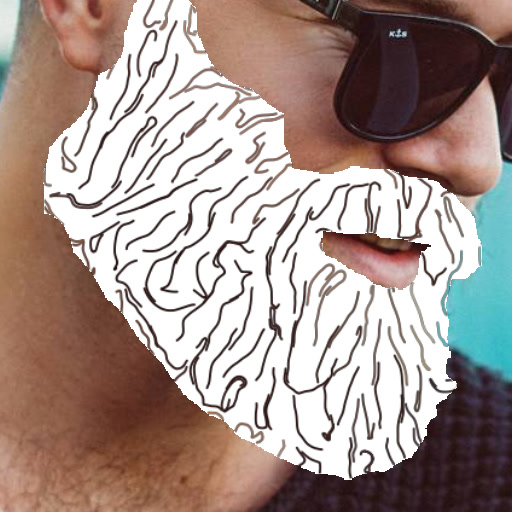}\\
            \includegraphics[width=0.1125\linewidth]{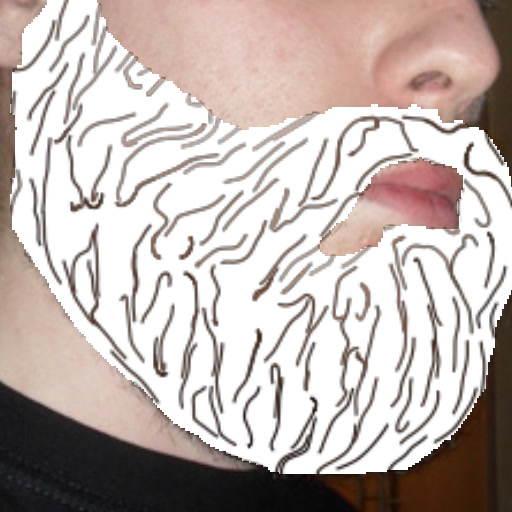}\\
\includegraphics[width=0.1125\linewidth]{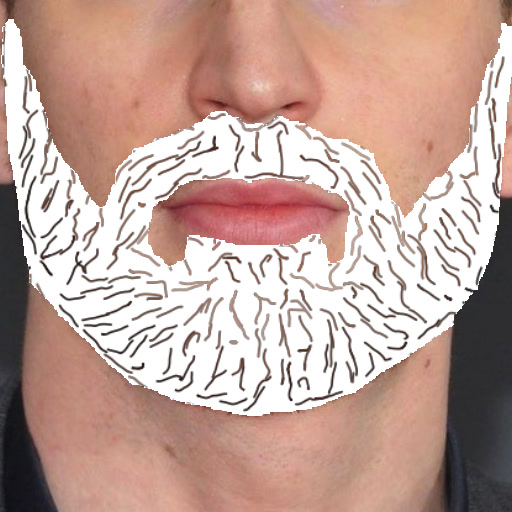}\\
            \includegraphics[width=0.1125\linewidth]{figs/raster/ablate/input/multistroke_stream_00_000059_img_01_composite.jpg}
\end{tabular}}\hspace{-10pt}
        \subfloat[Isola~\etal]{\begin{tabular}[b]{c}\includegraphics[width=0.1125\linewidth]{figs/raster/ablate/p2p/p2p_test_01_000012_img_01_output_full.jpg}\\        
            \includegraphics[width=0.1125\linewidth]{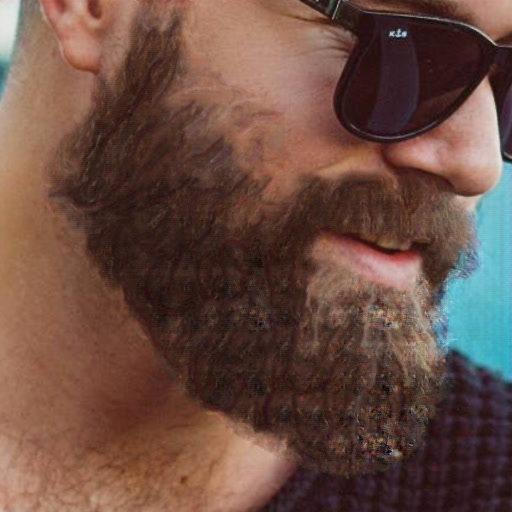}\\
            \includegraphics[width=0.1125\linewidth]{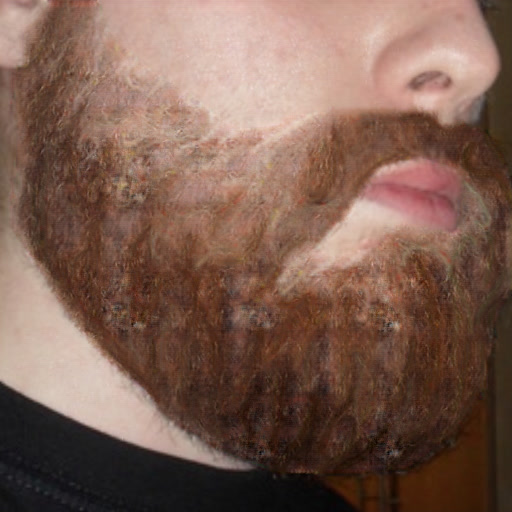}\\
\includegraphics[width=0.1125\linewidth]{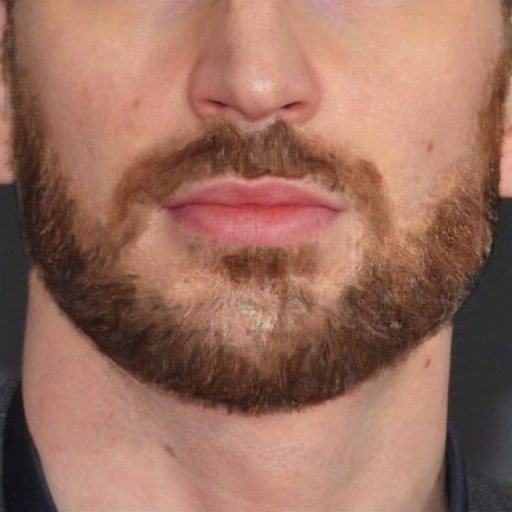}\\
            \includegraphics[width=0.1125\linewidth]{figs/raster/ablate/p2p/p2p_test_01_000059_img_01_output_full.jpg}
\end{tabular}}\hspace{-10pt}
        \subfloat[Single Network]{\begin{tabular}[b]{c}\includegraphics[width=0.1125\linewidth]{figs/raster/ablate/single/single_model_test_00_000012_img_01_output_full.jpg}\\        
            \includegraphics[width=0.1125\linewidth]{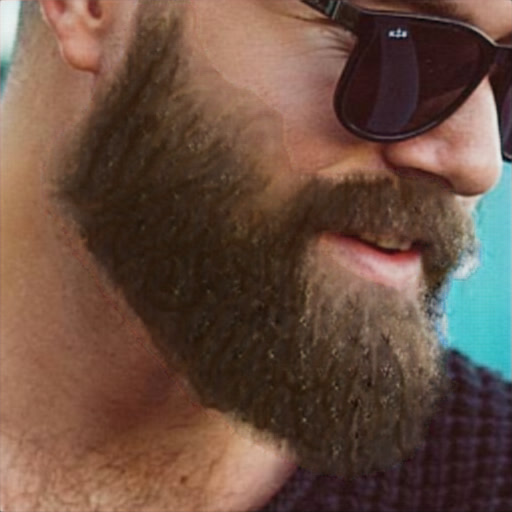}\\
            \includegraphics[width=0.1125\linewidth]{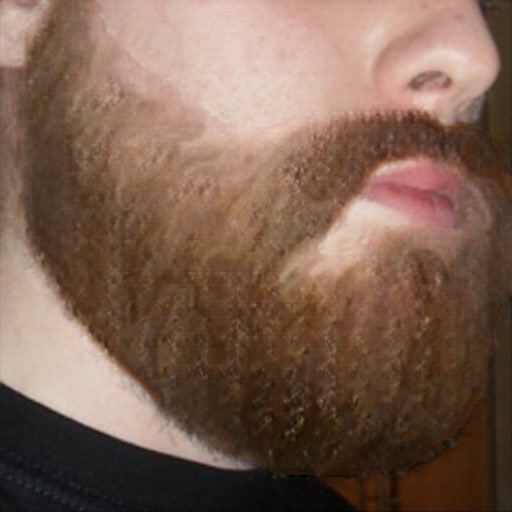}\\
\includegraphics[width=0.1125\linewidth]{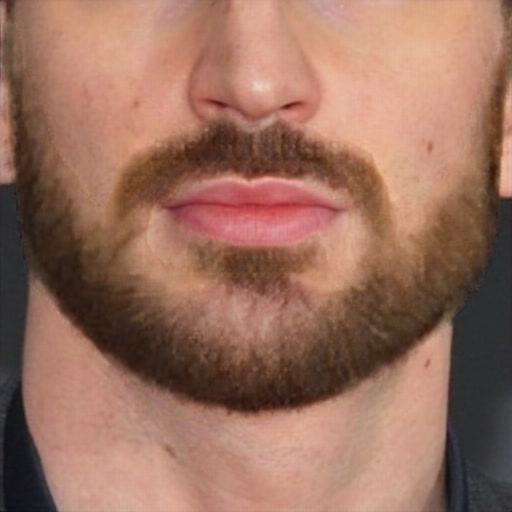}\\
            \includegraphics[width=0.1125\linewidth]{figs/raster/ablate/single/single_model_test_00_000059_img_01_output_full.jpg}
\end{tabular}}\hspace{-10pt}
        \subfloat[w/o GAN]{\begin{tabular}[b]{c}\includegraphics[width=0.1125\linewidth]{figs/raster/ablate/nogan/ablate_test_00_000012_img_01_output_full.jpg}\\        
            \includegraphics[width=0.1125\linewidth]{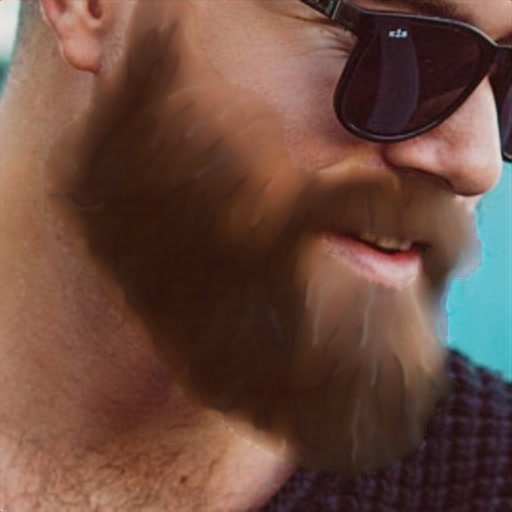}\\
            \includegraphics[width=0.1125\linewidth]{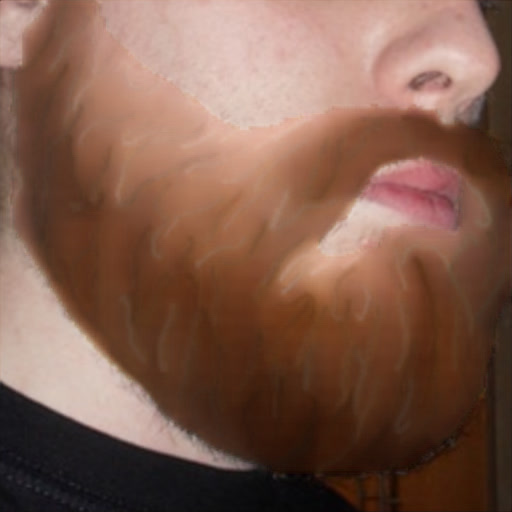}\\
\includegraphics[width=0.1125\linewidth]{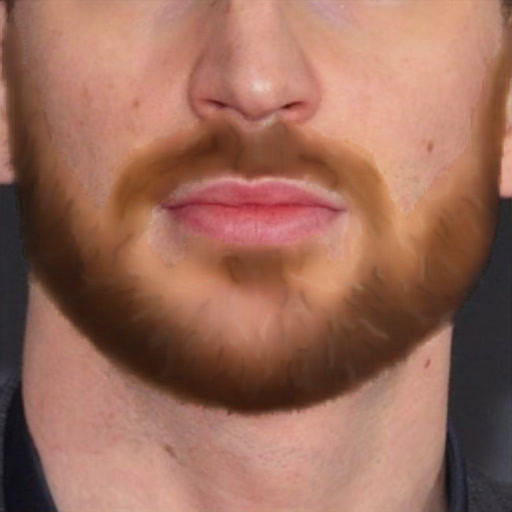}\\
            \includegraphics[width=0.1125\linewidth]{figs/raster/ablate/nogan/ablate_test_00_000059_img_01_output_full.jpg}
\end{tabular}}\hspace{-10pt}
        \subfloat[w/o VGG]{\begin{tabular}[b]{c}\includegraphics[width=0.1125\linewidth]{figs/raster/ablate/novgg/novgg_test_03_000012_img_01_output_full.jpg}\\        
            \includegraphics[width=0.1125\linewidth]{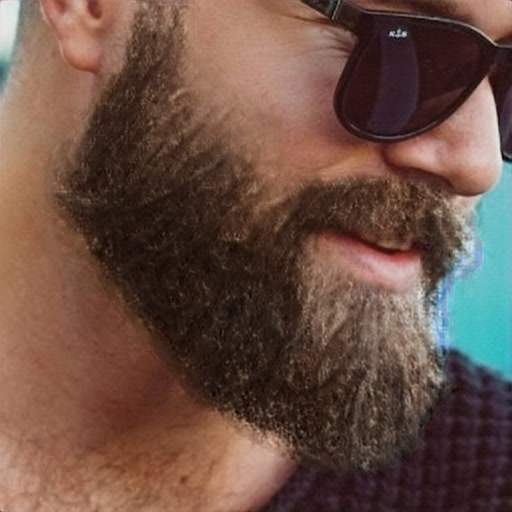}\\
            \includegraphics[width=0.1125\linewidth]{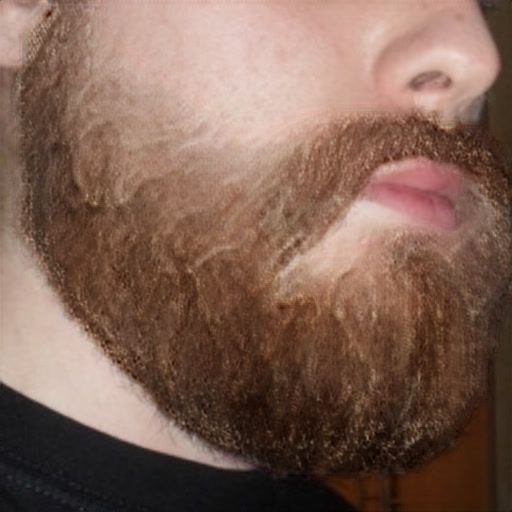}\\
\includegraphics[width=0.1125\linewidth]{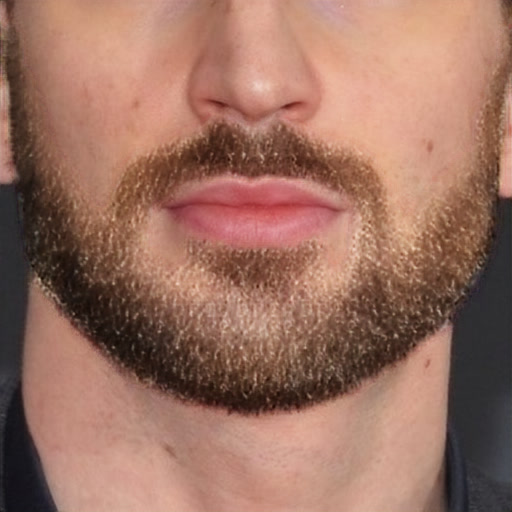}\\
            \includegraphics[width=0.1125\linewidth]{figs/raster/ablate/novgg/novgg_test_03_000059_img_01_output_full.jpg}
\end{tabular}}\hspace{-10pt}
        \subfloat[w/o synth. data]{\begin{tabular}[b]{c}\includegraphics[width=0.1125\linewidth]{figs/raster/ablate/no_synth_training/nosynth_test_00_000012_img_01_output_full.jpg}\\        
            \includegraphics[width=0.1125\linewidth]{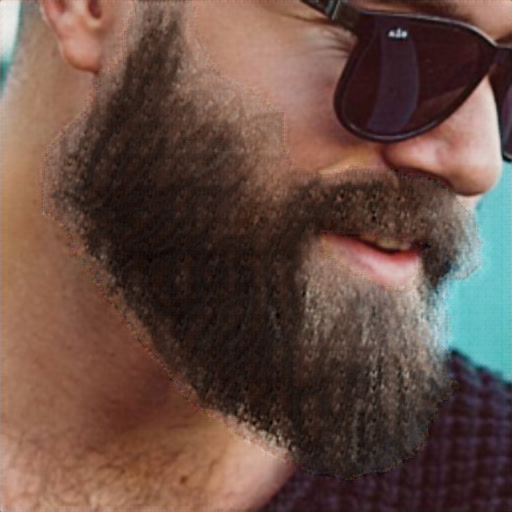}\\
            \includegraphics[width=0.1125\linewidth]{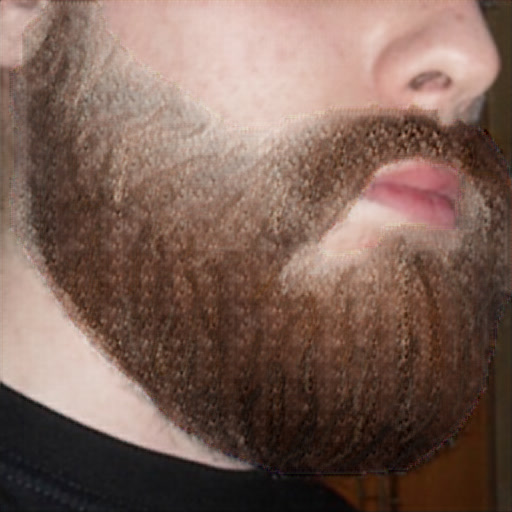}\\
\includegraphics[width=0.1125\linewidth]{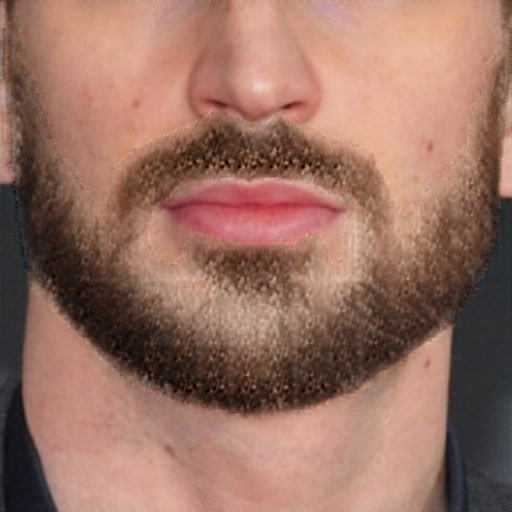}\\
            \includegraphics[width=0.1125\linewidth]{figs/raster/ablate/no_synth_training/nosynth_test_00_000059_img_01_output_full.jpg}
\end{tabular}}\hspace{-10pt}
        \subfloat[Ours, final]{\begin{tabular}[b]{c}\includegraphics[width=0.1125\linewidth]{figs/raster/ablate/ours/multistroke_test_11_000012_img_01_output_full.jpg}\\        
            \includegraphics[width=0.1125\linewidth]{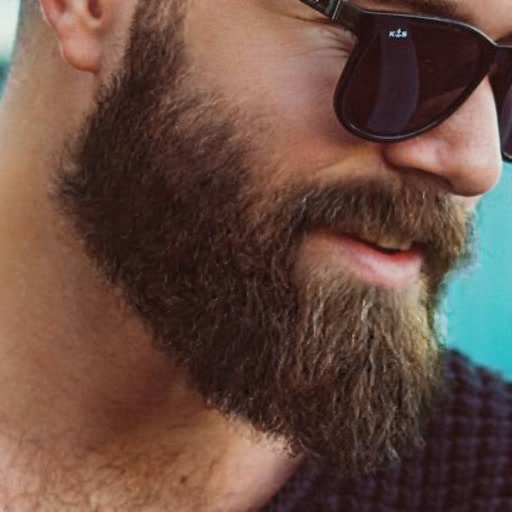}\\
            \includegraphics[width=0.1125\linewidth]{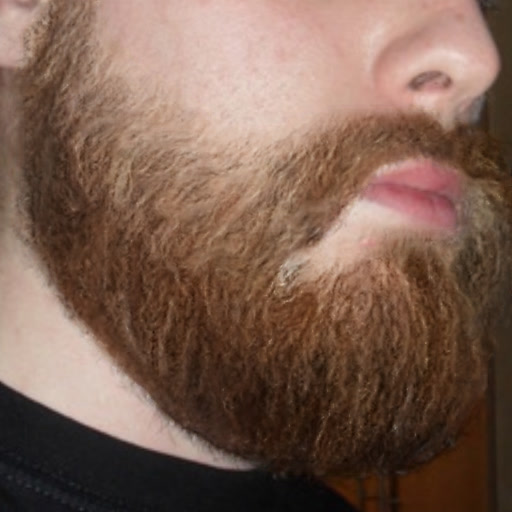}\\
\includegraphics[width=0.1125\linewidth]{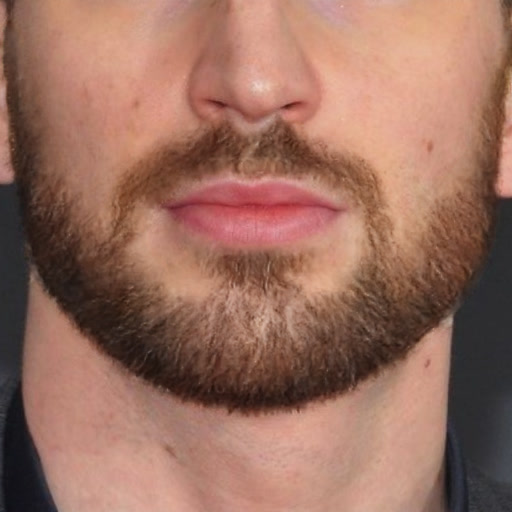}\\
            \includegraphics[width=0.1125\linewidth]{figs/raster/ablate/ours/multistroke_test_11_000059_img_01_output_full.jpg}
\end{tabular}}\hspace{-10pt}
        \subfloat[Ground truth]{\begin{tabular}[b]{c}\includegraphics[width=0.1125\linewidth]{figs/raster/ablate/gt/nosynth_test_00_000012_img_01_input_full.jpg}\\     
            \includegraphics[width=0.1125\linewidth]{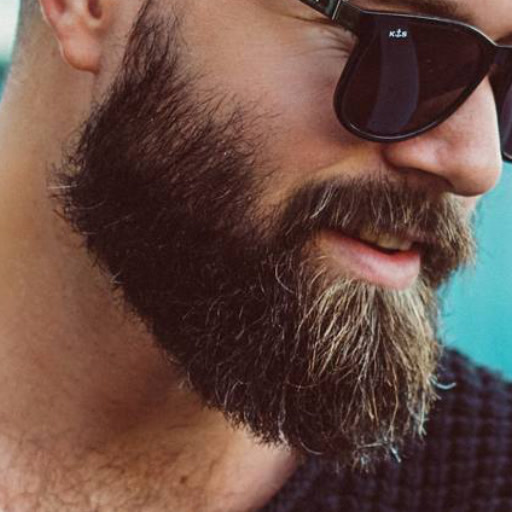}\\
            \includegraphics[width=0.1125\linewidth]{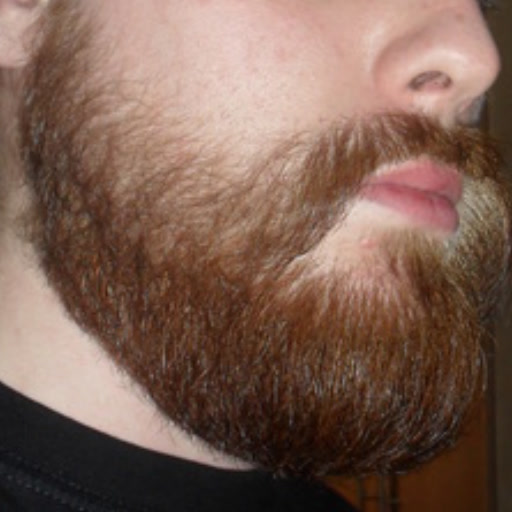}\\
\includegraphics[width=0.1125\linewidth]{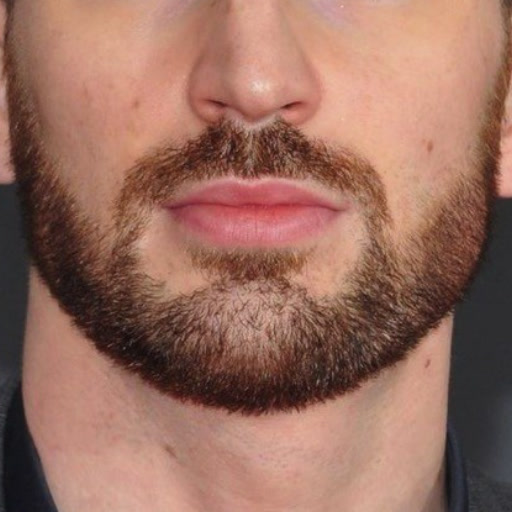}\\
            \includegraphics[width=0.1125\linewidth]{figs/raster/ablate/gt/nosynth_test_00_000059_img_01_input_full.jpg}
\end{tabular}}\hfill
        \caption{Qualitative comparisons for our ablation study.}
  {}
  \label{fig:supp_ablation}
\end{figure*}

We show additional selected qualitative results from the ablation study described in Sec.~\ref{sec:results} in Fig.~\ref{fig:supp_ablation}. With the addition of the refinement network, our results contain more subtle details and have fewer artifacts at skin boundaries than when using only one network. Adversarial loss adds more fine-scale details, while VGG perceptual loss substantially reduces noisy artifacts. Compared with networks trained with only real image data, our final result has clearer definition for individual hair strands and has a higher dynamic range, which is preferable if users are to perform elaborate image editing. With all these components, our networks produce results of much higher quality than the baseline network of \cite{isola2016image}.

\section{User study}
\label{sec:user_study}
\input{uist_submission/user_study.tex}

\section{Implementation Details}
\label{sec:implementation}
\input{uist_submission/implementation}

%% file: uist_submission/ui.tex
\begin{figure}[h!]
\includegraphics[width=\columnwidth]{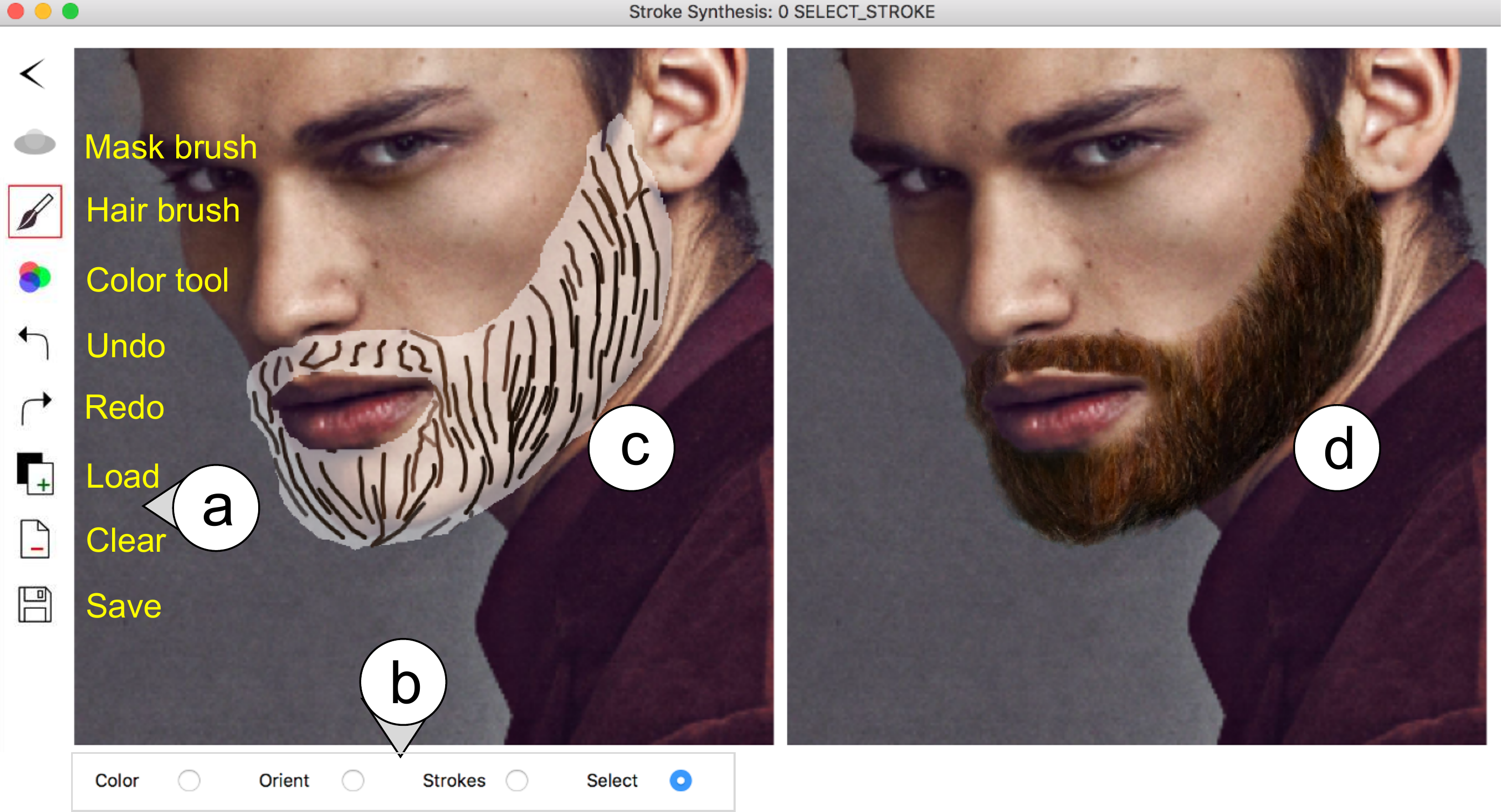}
  \caption{
  The user interface, consisting of global (a) and contextual (b) toolbars, and input (c) and result preview (d) canvases.}
  \label{fig:interface}
\end{figure}
\begin{figure}[h!]
\includegraphics[width=\columnwidth]{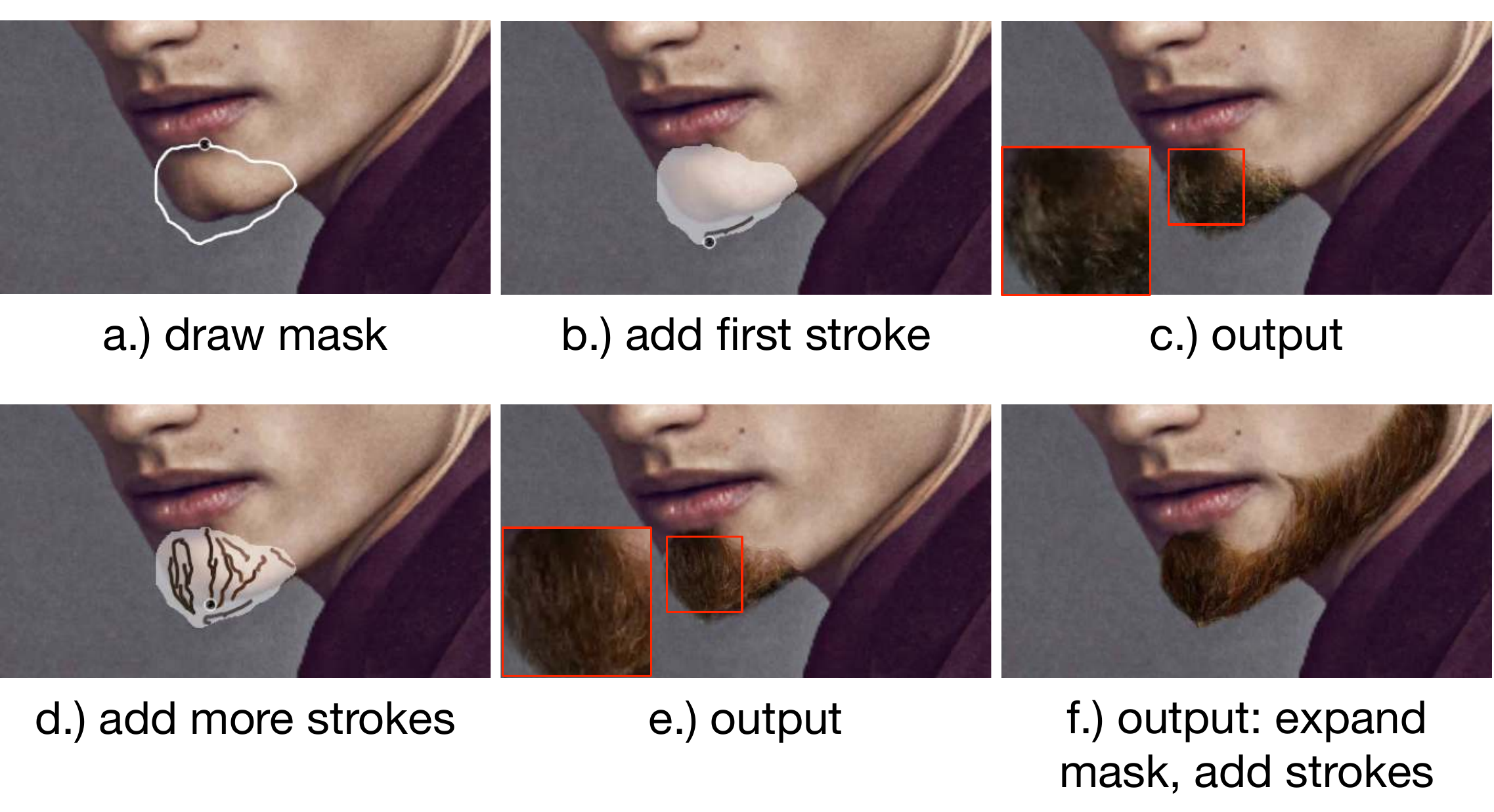}
  \vspace{-7mm}
  \caption{
  An example of interactive facial hair creation from scratch on a clean-shaven target image. After creating the initial mask (a), the user draws strokes in this region that define the local color and shape of the hair (b-e). While a single stroke (b) has little effect on much of the masked region (c), adding more strokes (d) results in more control over the output (e). The user can also change the mask shape or add more strokes (f) to adjust the overall hairstyle.}
  \vspace{2mm}
  \label{fig:example_workflow}
\end{figure}
\begin{figure}[h!]
\vspace{-1mm}
\includegraphics[width=\columnwidth]{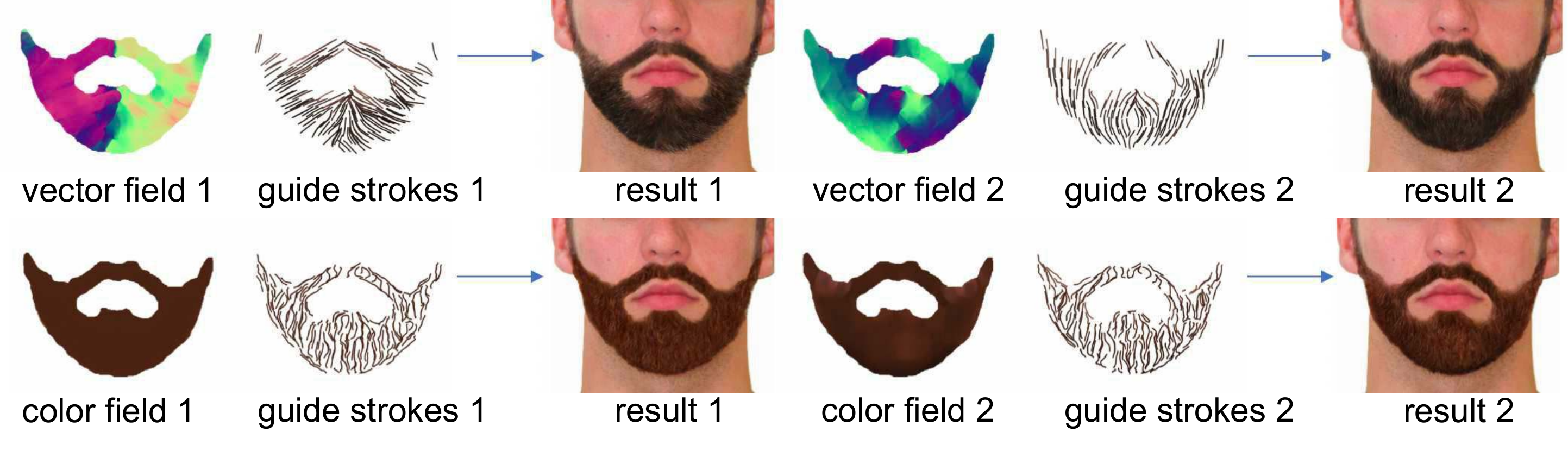}
\caption{Changes to the vector and color fields cause corresponding changes in the guide strokes, and also the final synthesized results.
\label{fig:field_changes}}
\vspace{-2mm}
\end{figure}

Fig.~\ref{fig:interface} shows the interface of our system. As described in the paper (Sec. 6), we provide tools for synthesizing a coarse initialization of the target hairstyle given only the user-drawn mask and selected color; separately manipulating the color and vector fields used to automatically extract guide strokes of the appropriate shape and color from this initial estimate; and drawing, removing, and changing individual strokes to make local edits to the final synthesized image.
See Fig.~\ref{fig:example_workflow} for an example of iterative refinement of an image using our provided input tools for mask creation and individual stroke drawing with the corresponding output, and Fig.~\ref{fig:field_changes} for an example of vector/color field editing, that easily changes the structure and color of the guide strokes. Also see Fig. 4 in the paper and the example sessions in the supplementary video for examples of conditional inpainting and other editing operations. 

\begin{figure}[h!]
\includegraphics[width=\columnwidth]{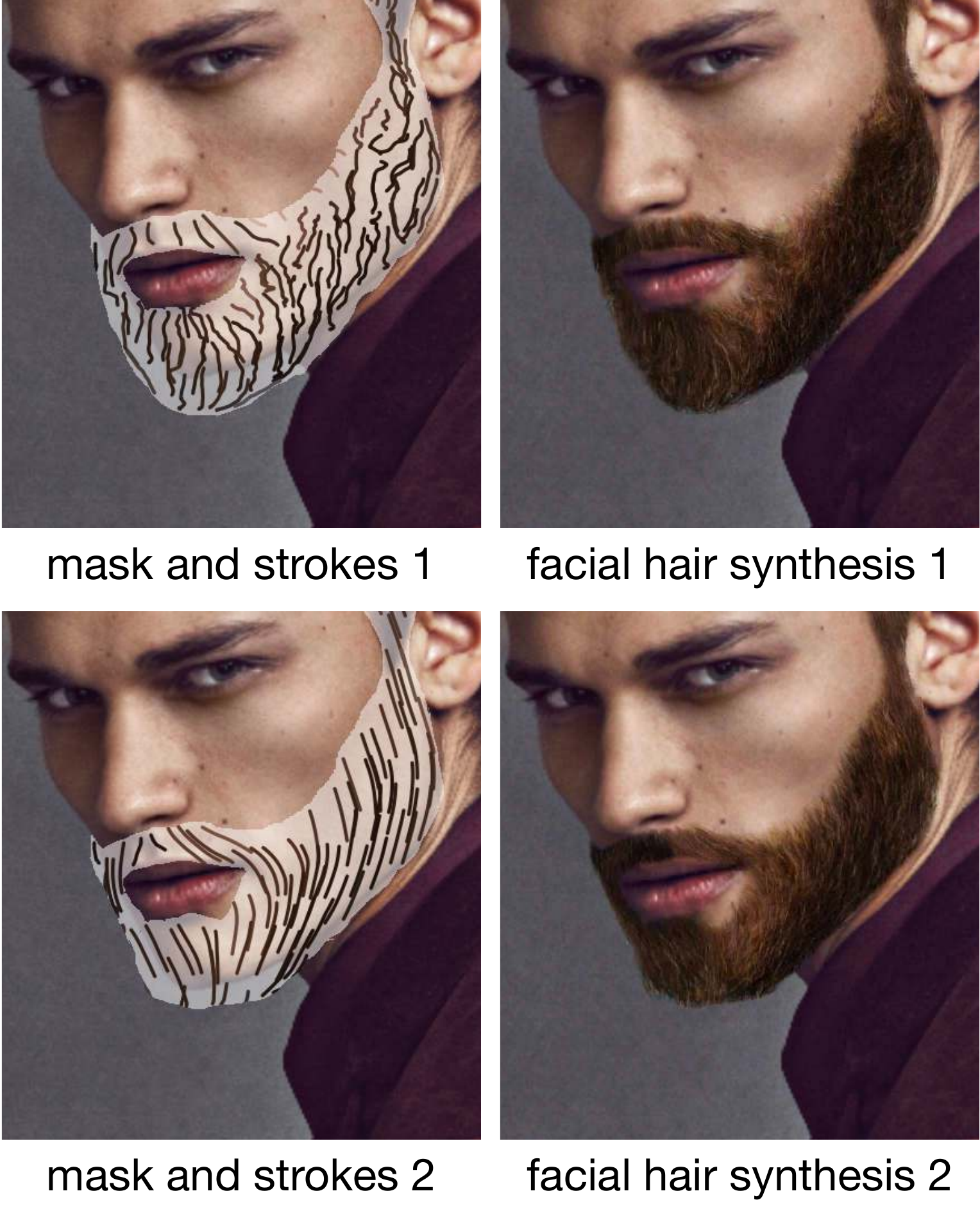}
\vspace{-6mm}
\caption{Subtle changes to the structure and appearance of the strokes used for final image synthesis cause corresponding changes in the final synthesized result.
\label{fig:stroke_changes}}
\vspace{-3mm}
\end{figure}

Fig.~\ref{fig:stroke_changes} shows an example of how the overall structure of a synthesized hairstyle can be changed by making adjustments to the structure of the user-provided guide strokes. By using strokes with the overall colors of those in row 1, column 1, but with different shapes, such as the smoother and more coherent strokes as in row 2, column 1, we can generate a correspondingly smooth and coherent hairstyle, (row 2, column 2).

\vspace{-3mm}

\paragraph{Facial hair reference database}
We provide a library of sample images from our facial hair database that can be used as visual references for target hairstyles. The user can select colors from regions of these images for the initial color mask, individual strokes and brush-based color field editing. This allows users to easily choose colors that represent the overall and local appearance of the desired hairstyle. Users may also copy and paste selected strokes from these images directly into the target region, so as to directly emulate the appearance of the reference image. This can be done using either the color of these selected strokes in the reference image, or merely their shape with the selected color and transparency settings so as to emulate the local structure of the reference image within the global structure and appearance of the target image.

%% file: uist_submission/user_study.tex
We conducted a preliminary user study to evaluate the usability of our system. The study included 8 users. 1 user was a professional technical artist, while the others were non-professionals, including novices with minimal to moderate prior experience with technical drawing or image editing. When asked to rate their prior experience as a technical artist on a scale of $1-5$, with $1$ indicating no prior experience and $5$ indicating a professionally trained technical artist, the average score was $3.19$.

\paragraph{Procedure}
The study consisted of three sessions: a warm-up session (approximately 10 min), a target session (15-25 min), and an open session (10 min). The users were then asked to provide feedback by answering a set of questions to quantitatively and qualitatively evaluate their experience. In the warm-up session, users were introduced to the input operations and workflow and then asked to familiarize themselves with these tools by synthesizing facial hair on a clean-shaven source image similar to that seen in a reference image. For the target session, the participants were given a new reference portrait image and asked to create similar hair on a different clean-shaven subject via (1) our interface, and (2) Brushables~\cite{Lukac:2015:BEE}. For Brushables, the user was asked to draw an vector field corresponding to the overall shape and orientation of the facial hairstyle in the target image. A patch of facial hair taken directly from the target image was used with this input to automatically synthesize the output facial hair.
For the open session, we let the participants explore the full functionality of our system and uncover potential usability issues by creating facial hair with arbitrary structures and colors.

\begin{figure}[h!]
  \centering
\includegraphics[width=0.75\columnwidth]{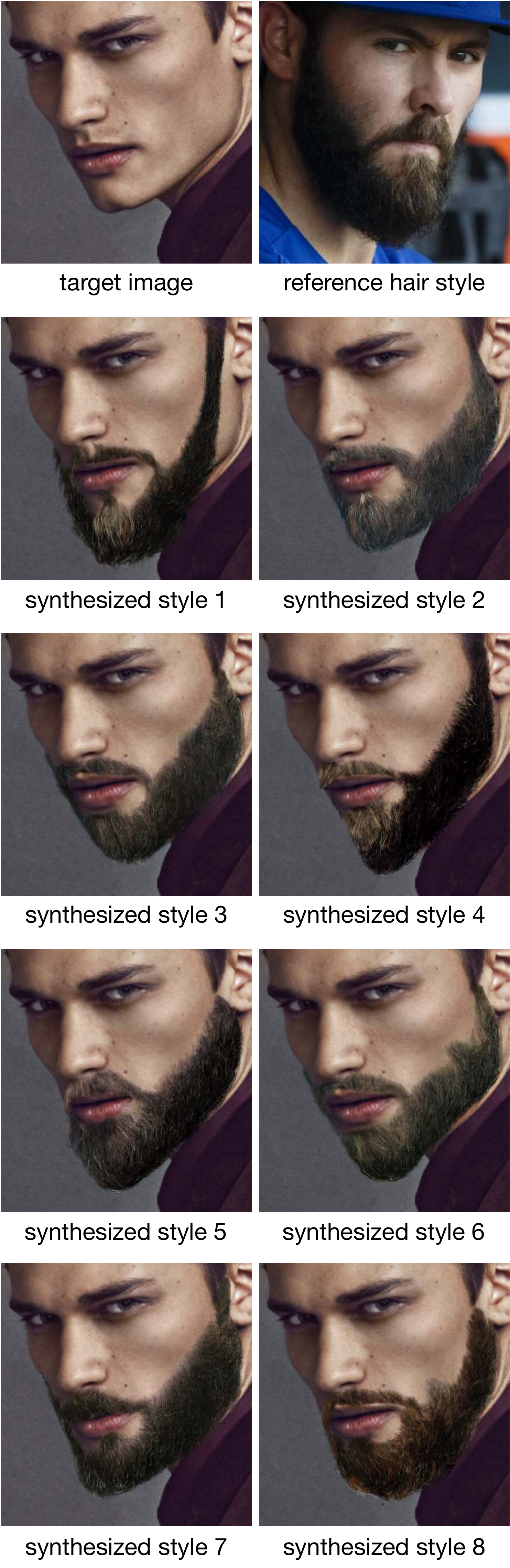}
  \caption{
  Example results generated by participants in our user study, when asked to synthesize a style resembling the target image (top left). No user had prior experience using our interface. Subjects took an average of 14 minutes (and no more than 19 minutes) to create the portrayed images.}
  \label{fig:study_results}
\end{figure}

\paragraph{Outcome}

Fig.~\ref{fig:study_results} provides qualitative results from the target session. Row 2, column depicts the result from the professional artist, while other results are from non-professionals with no artistic training. The users took between $6-19$ minutes to create the target image using our tool. The average session time was $14$ minutes. The users required an average of 116 strokes to synthesize the target facial hairstyle on the source subject. 39 of these strokes were required to draw and edit the initial mask defining the region in which synthesis is performed. The remaining 77 were brush strokes used to edit the color and vector fields used to automatically generate strokes in this region, and to draw the individual strokes used to perform the final refinement. On average a user performed 19 brush strokes to edit the vector field, 17 brush strokes to edit the color field, and drew 41 individual strokes. We note that these numbers include individual strokes deleted by the user if their impact on the resulting image was deemed unsatisfactory. Overall, these numbers indicate, as does the provided feedback, that the color and brush editing tools were useful in reducing the number of individual strokes required to synthesize the final image. We use the original Brushables implementation for comparison, which does not provide statistics on the number of operations performed by the user, and as such we can not report these statistics for the Brushables session.

\paragraph{Feedback}

We asked the users to rate our method in terms of ease-of-use and their perceived quality of the final image they created during the target session. On a scale of $1-5$ (higher is better), the users rated our system $3.6$ in terms of ease of use, and $4.06$ in terms of their synthesized result matching the target facial hairstyle. When asked to measure how satisfied they were with the result given the amount of time they spent creating it and becoming familiar with the system, the average score was $4.0$. Furthermore, $100\%$ of the users preferred our system over Brushables for the task of facial hair editing.

After being introduced to its interface, users spent between $3-5$ minutes ($4$ minutes on average) working with Brushables to attempt to synthesize the target hairstyle. While less time was required to synthesize the results with Brushables, the users generally found the results achieved by copying regions of the source texture sample directly into the specified target region to be very unsatisfactory. By simply attempting to create an vector field roughly matching the structure of the target hairstyle and then synthesizing the result, users had little control over the subtle local details necessary to synthesize a plausible result. Furthermore, as Brushables required approximately 30 seconds to synthesize the entire facial hairstyle given the complete user-defined vector field, iterative experimentation was far more difficult than when using our approach, which allows for immediately visualizing the results of minor editing operations. Thus, users chose not to experiment with Brushables long enough to produce more satisfactory results.

Overall, the participants found our system novel and useful. When asked what features they found most useful, some users commented that they liked the ability to create a rough approximation of the target hairstyle given only a mask and average color. Others strongly appreciated the color and orientation brushes, as these allowed them to separately change the color and structure of the initial estimate, and to change large regions of the image without drawing each individual stroke with the appropriate shape and color. In contrast, drawing each individual stroke manually was not perceived as especially useful, as it required significantly more effort to experiment with creating and removing individual strokes to produce a combination with the appropriate shape and color to achieve the desired result. However, overall participants were able to achieve reasonable results such as those in Fig.~\ref{fig:study_results} primarily relying on the color and vector field brushes to edit the initial synthesis results produced when selecting the mask shape and color. Relatively few individual strokes were ultimately required to refine the results.

This feedback suggests that the increasing level of granularity enabled by our system (creating a rough initial estimate, modifying the local shape and color, and then refining small details with a few individual strokes) is an effective approach. Furthermore, the participants reported that the real-time synthesis of the generated image allowed for intuitive iterative refinement, which provided helpful visual guidance crucial for producing satisfactory results.

During the open session at the end, users enjoyed experimenting with our tools to creatively generate unconventional hairstyles with unusual shapes and colors. However, as many of these styles were well outside the range of natural shapes and colors seen in the images used to train our system, the results were less realistic than those constrained to resemble a more conventional hairstyle.

%% file: uist_submission/implementation.tex
\paragraph{Network Training.} We train the first network in our two-stage pipeline first with synthetic then with real data. Then, we train both stages in an end-to-end manner with real images while keeping the losses for both stages. 

When training the first network individually, we use an initial learning rate of 0.0002 and momentum of 0.5. The learning rate is halved twice during this training process such that in the final epochs the learning rate is reduced to 0.00005.
During end-to-end training of both networks, the initial learning rate is reduced to 0.0001 and a momentum of 0.75 is used. As before, the learning rate is halved twice during training, resulting in a final learning rate of 0.000025.

Both architectures are fully convolutional and thus can take input images of any resolution. However, we scale all our training data to a resolution of $512 \times 512$. We train both networks via the Adam optimizer \cite{DBLP:journals/corr/KingmaB14} on an NVIDIA Titan X GPU using the Torch framework \cite{Collobert02torch:a}. We first train each stage of the networks for 50 epochs, which takes about 24 hours. Then refine the first network using real image data to train for 25 epochs, which takes about 12 hours.

\paragraph{Runtime Performance.} The user interface is designed to allow for input using either a traditional mouse for novice users, or the tablet and stylus tools used by digital artists. For run-time interaction, passing one image through our network takes a total of 600 milliseconds. We transmit the input image to a server running our networks. The total time between making an update to the input and seeing the corresponding on average thus takes roughly 1.5 seconds. The vector field used for the initial stroke extraction is also performed using CUDA for GPU acceleration, and takes roughly 120 milliseconds for a $512 \times 512$ image. 

\section{User Interaction Time}
Each interactively generated result seen in the figures in this paper (with the exception of some examples from the user study described below) was made in no more than 15 minutes by a user with no artistic training. On average, the depicted examples in this paper took between 2-6 minutes, depending on the size and complexity of the hairstyle. Simple examples, such as the short hairstyle in Fig. 1. in the main paper can be created in 2-3 minutes, while more complex styles, such as the sparse beard in Fig. 7 (row 2, column 7) in the main paper, took 10-15 minutes. We refer to the supplementary video for a live recording of several editing sessions.

%% file: proceedings.bbl
\begin{thebibliography}{10}\itemsep=-1pt

\bibitem{barnes2009patchmatch}
Connelly Barnes, Eli Shechtman, Adam Finkelstein, and Dan~B Goldman.
\newblock Patchmatch: A randomized correspondence algorithm for structural
  image editing.
\newblock {\em ACM Transactions on Graphics-TOG}, 28(3):24, 2009.

\bibitem{barnes2017survey}
Connelly Barnes and Fang-Lue Zhang.
\newblock A survey of the state-of-the-art in patch-based synthesis.
\newblock {\em Computational Visual Media}, 3(1):3--20, 2017.

\bibitem{beeler2012coupled}
Thabo Beeler, Bernd Bickel, Gioacchino Noris, Paul Beardsley, Steve Marschner,
  Robert~W Sumner, and Markus Gross.
\newblock Coupled 3d reconstruction of sparse facial hair and skin.
\newblock {\em ACM Transactions on Graphics (ToG)}, 31(4):117, 2012.

\bibitem{bertalmio2000image}
Marcelo Bertalmio, Guillermo Sapiro, Vincent Caselles, and Coloma Ballester.
\newblock Image inpainting.
\newblock In {\em Proceedings of the 27th annual conference on Computer
  graphics and interactive techniques}, pages 417--424. ACM
  Press/Addison-Wesley Publishing Co., 2000.

\bibitem{bertalmio2003simultaneous}
Marcelo Bertalmio, Luminita Vese, Guillermo Sapiro, and Stanley Osher.
\newblock Simultaneous structure and texture image inpainting.
\newblock {\em IEEE transactions on image processing}, 12(8):882--889, 2003.

\bibitem{brock2016neural}
Andrew Brock, Theodore Lim, James~M Ritchie, and Nick Weston.
\newblock Neural photo editing with introspective adversarial networks.
\newblock {\em arXiv preprint arXiv:1609.07093}, 2016.

\bibitem{chai2016autohair}
Menglei Chai, Tianjia Shao, Hongzhi Wu, Yanlin Weng, and Kun Zhou.
\newblock Autohair: Fully automatic hair modeling from a single image.
\newblock {\em ACM Transactions on Graphics (TOG)}, 35(4):116, 2016.

\bibitem{Chai:2013:DHM}
Menglei Chai, Lvdi Wang, Yanlin Weng, Xiaogang Jin, and Kun Zhou.
\newblock Dynamic hair manipulation in images and videos.
\newblock {\em ACM Trans. Graph.}, 32(4):75:1--75:8, July 2013.

\bibitem{chai2012single}
Menglei Chai, Lvdi Wang, Yanlin Weng, Yizhou Yu, Baining Guo, and Kun Zhou.
\newblock Single-view hair modeling for portrait manipulation.
\newblock {\em ACM Transactions on Graphics (TOG)}, 31(4):116, 2012.

\bibitem{chang2018pairedcyclegan}
Huiwen Chang, Jingwan Lu, Fisher Yu, and Adam Finkelstein.
\newblock Pairedcyclegan: Asymmetric style transfer for applying and removing
  makeup.
\newblock In {\em 2018 IEEE Conference on Computer Vision and Pattern
  Recognition (CVPR)}, 2018.

\bibitem{choe2005statistical}
Byoungwon Choe and Hyeong-Seok Ko.
\newblock A statistical wisp model and pseudophysical approaches for
  interactive hairstyle generation.
\newblock {\em IEEE Transactions on Visualization and Computer Graphics},
  11(2):160--170, 2005.

\bibitem{Collobert02torch:a}
Ronan Collobert, Samy Bengio, and Johnny Marithoz.
\newblock Torch: A modular machine learning software library, 2002.

\bibitem{criminisi2004region}
Antonio Criminisi, Patrick P{\'e}rez, and Kentaro Toyama.
\newblock Region filling and object removal by exemplar-based image inpainting.
\newblock {\em IEEE Transactions on image processing}, 13(9):1200--1212, 2004.

\bibitem{darabi2012image}
Soheil Darabi, Eli Shechtman, Connelly Barnes, Dan~B Goldman, and Pradeep Sen.
\newblock Image melding: Combining inconsistent images using patch-based
  synthesis.
\newblock {\em ACM Trans. Graph.}, 31(4):82--1, 2012.

\bibitem{daz3d}
{Daz Productions}, 2017.
\newblock \url{https://www.daz3d.com/}.

\bibitem{dekel2018sparse}
Tali Dekel, Chuang Gan, Dilip Krishnan, Ce Liu, and William~T Freeman.
\newblock Sparse, smart contours to represent and edit images.
\newblock In {\em Proceedings of the IEEE Conference on Computer Vision and
  Pattern Recognition}, pages 3511--3520, 2018.

\bibitem{denton2015deep}
Emily~L Denton, Soumith Chintala, Rob Fergus, et~al.
\newblock Deep generative image models using a laplacian pyramid of adversarial
  networks.
\newblock In {\em Advances in neural information processing systems}, pages
  1486--1494, 2015.

\bibitem{dolhansky2017eye}
Brian Dolhansky and Cristian~Canton Ferrer.
\newblock Eye in-painting with exemplar generative adversarial networks.
\newblock {\em arXiv preprint arXiv:1712.03999}, 2017.

\bibitem{Efros:2001:IQT}
Alexei~A. Efros and William~T. Freeman.
\newblock Image quilting for texture synthesis and transfer.
\newblock In {\em Proceedings of the 28th Annual Conference on Computer
  Graphics and Interactive Techniques}, SIGGRAPH '01, pages 341--346. ACM,
  2001.

\bibitem{Efros:1999:TSN}
Alexei~A. Efros and Thomas~K. Leung.
\newblock Texture synthesis by non-parametric sampling.
\newblock In {\em IEEE ICCV}, pages 1033--, 1999.

\bibitem{fivser2016stylit}
Jakub Fi{\v{s}}er, Ond{\v{r}}ej Jamri{\v{s}}ka, Michal Luk{\'a}{\v{c}}, Eli
  Shechtman, Paul Asente, Jingwan Lu, and Daniel S{\`y}kora.
\newblock Stylit: illumination-guided example-based stylization of 3d
  renderings.
\newblock {\em ACM Transactions on Graphics (TOG)}, 35(4):92, 2016.

\bibitem{Fiser:2017:ESS}
Jakub Fi\v{s}er, Ond\v{r}ej Jamri\v{s}ka, David Simons, Eli Shechtman, Jingwan
  Lu, Paul Asente, Michal Luk\'{a}\v{c}, and Daniel S\'{y}kora.
\newblock Example-based synthesis of stylized facial animations.
\newblock {\em ACM Trans. Graph.}, 36(4):155:1--155:11, July 2017.

\bibitem{Gatys:2015:ANA}
Leon~A. Gatys, Alexander~S. Ecker, and Matthias Bethge.
\newblock A neural algorithm of artistic style.
\newblock {\em CoRR}, abs/1508.06576, 2015.

\bibitem{Gatys:2015:TSU}
Leon~A. Gatys, Alexander~S. Ecker, and Matthias Bethge.
\newblock Texture synthesis using convolutional neural networks.
\newblock In {\em Proceedings of the 28th International Conference on Neural
  Information Processing Systems}, NIPS'15, pages 262--270, Cambridge, MA, USA,
  2015. MIT Press.

\bibitem{gatys2016image}
Leon~A Gatys, Alexander~S Ecker, and Matthias Bethge.
\newblock Image style transfer using convolutional neural networks.
\newblock In {\em Proc. CVPR}, pages 2414--2423, 2016.

\bibitem{goodfellow2014generative}
Ian Goodfellow, Jean Pouget-Abadie, Mehdi Mirza, Bing Xu, David Warde-Farley,
  Sherjil Ozair, Aaron Courville, and Yoshua Bengio.
\newblock Generative adversarial nets.
\newblock In {\em Advances in neural information processing systems}, pages
  2672--2680, 2014.

\bibitem{han2006fast}
Jianwei Han, Kun Zhou, Li-Yi Wei, Minmin Gong, Hujun Bao, Xinming Zhang, and
  Baining Guo.
\newblock Fast example-based surface texture synthesis via discrete
  optimization.
\newblock {\em The Visual Computer}, 22(9-11):918--925, 2006.

\bibitem{DBLP:journals/corr/HeuselRUNKH17}
Martin Heusel, Hubert Ramsauer, Thomas Unterthiner, Bernhard Nessler,
  G{\"{u}}nter Klambauer, and Sepp Hochreiter.
\newblock Gans trained by a two time-scale update rule converge to a nash
  equilibrium.
\newblock {\em CoRR}, abs/1706.08500, 2017.

\bibitem{hu2014robust}
Liwen Hu, Chongyang Ma, Linjie Luo, and Hao Li.
\newblock Robust hair capture using simulated examples.
\newblock {\em ACM Transactions on Graphics (TOG)}, 33(4):126, 2014.

\bibitem{Hu:2015:SHM}
Liwen Hu, Chongyang Ma, Linjie Luo, and Hao Li.
\newblock Single-view hair modeling using a hairstyle database.
\newblock {\em ACM Trans. Graph.}, 34(4):125:1--125:9, July 2015.

\bibitem{hu2017avatar}
Liwen Hu, Shunsuke Saito, Lingyu Wei, Koki Nagano, Jaewoo Seo, Jens Fursund,
  Iman Sadeghi, Carrie Sun, Yen-Chun Chen, and Hao Li.
\newblock Avatar digitization from a single image for real-time rendering.
\newblock {\em ACM Transactions on Graphics (TOG)}, 36(6):195, 2017.

\bibitem{isola2016image}
Phillip Isola, Jun-Yan Zhu, Tinghui Zhou, and Alexei~A Efros.
\newblock Image-to-image translation with conditional adversarial networks.
\newblock {\em arXiv preprint arXiv:1611.07004}, 2016.

\bibitem{DBLP:journals/corr/abs-1902-06838}
Youngjoo Jo and Jongyoul Park.
\newblock {SC-FEGAN:} face editing generative adversarial network with user's
  sketch and color.
\newblock {\em CoRR}, abs/1902.06838, 2019.

\bibitem{Johnson:2016:PLR}
Justin Johnson, Alexandre Alahi, and Fei{-}Fei Li.
\newblock Perceptual losses for real-time style transfer and super-resolution.
\newblock {\em CoRR}, abs/1603.08155, 2016.

\bibitem{DBLP:journals/corr/JohnsonAL16}
Justin Johnson, Alexandre Alahi, and Fei{-}Fei Li.
\newblock Perceptual losses for real-time style transfer and super-resolution.
\newblock {\em CoRR}, abs/1603.08155, 2016.

\bibitem{kaspar2015self}
Alexandre Kaspar, Boris Neubert, Dani Lischinski, Mark Pauly, and Johannes
  Kopf.
\newblock Self tuning texture optimization.
\newblock {\em Computer Graphics Forum}, 34(2):349--359, 2015.

\bibitem{DBLP:journals/corr/KellerWRSAKK17}
Alexander Keller, Carsten W{\"{a}}chter, Matthias Raab, Daniel Seibert, Dietger
  van Antwerpen, Johann Kornd{\"{o}}rfer, and Lutz Kettner.
\newblock The iray light transport simulation and rendering system.
\newblock {\em CoRR}, abs/1705.01263, 2017.

\bibitem{kemelmacher2016transfiguring}
Ira Kemelmacher-Shlizerman.
\newblock Transfiguring portraits.
\newblock {\em ACM Transactions on Graphics (TOG)}, 35(4):94, 2016.

\bibitem{kim2002interactive}
Tae-Yong Kim and Ulrich Neumann.
\newblock Interactive multiresolution hair modeling and editing.
\newblock {\em ACM Transactions on Graphics (TOG)}, 21(3):620--629, 2002.

\bibitem{DBLP:journals/corr/KingmaB14}
Diederik~P. Kingma and Jimmy Ba.
\newblock Adam: {A} method for stochastic optimization.
\newblock {\em CoRR}, abs/1412.6980, 2014.

\bibitem{Kwatra:2005:TOE}
Vivek Kwatra, Irfan Essa, Aaron Bobick, and Nipun Kwatra.
\newblock Texture optimization for example-based synthesis.
\newblock {\em ACM Trans. Graph.}, 24(3):795--802, July 2005.

\bibitem{Kwatra:2003:GTI}
Vivek Kwatra, Arno Sch\"{o}dl, Irfan Essa, Greg Turk, and Aaron Bobick.
\newblock Graphcut textures: Image and video synthesis using graph cuts.
\newblock In {\em Proc. SIGGRAPH}, SIGGRAPH '03, pages 277--286. ACM, 2003.

\bibitem{Kyprianidis:2011:CEF}
Jan~Eric Kyprianidis and Henry Kang.
\newblock Image and video abstraction by coherence-enhancing filtering.
\newblock {\em Computer Graphics Forum}, 30(2):593--–602, 2011.
\newblock Proceedings Eurographics 2011.

\bibitem{lasram2012parallel}
Anass Lasram and Sylvain Lefebvre.
\newblock Parallel patch-based texture synthesis.
\newblock In {\em Proceedings of the Fourth ACM SIGGRAPH/Eurographics
  conference on High-Performance Graphics}, pages 115--124. Eurographics
  Association, 2012.

\bibitem{daz3dwhiskers}
{Laticis Imagery}, 2017.
\newblock \url{https://www.daz3d.com/whiskers-for-genesis-3-male-s}.

\bibitem{lefebvre2006appearance}
Sylvain Lefebvre and Hugues Hoppe.
\newblock Appearance-space texture synthesis.
\newblock {\em ACM Trans. Graph.}, 25(3):541--548, 2006.

\bibitem{Li:2016:PRT}
Chuan Li and Michael Wand.
\newblock Precomputed real-time texture synthesis with markovian generative
  adversarial networks.
\newblock {\em CoRR}, abs/1604.04382, 2016.

\bibitem{Liao:2017:VAT}
Jing Liao, Yuan Yao, Lu Yuan, Gang Hua, and Sing~Bing Kang.
\newblock Visual attribute transfer through deep image analogy.
\newblock {\em ACM Trans. Graph.}, 36(4):120:1--120:15, July 2017.

\bibitem{Lukac:2015:BEE}
Michal Luk\'{a}\v{c}, Jakub Fiser, Paul Asente, Jingwan Lu, Eli Shechtman, and
  Daniel S\'{y}kora.
\newblock Brushables: Example-based edge-aware directional texture painting.
\newblock {\em Comput. Graph. Forum}, 34(7):257--267, 2015.

\bibitem{Lukac:2013:PFT}
Michal Luk\'{a}\v{c}, Jakub Fi\v{s}er, Jean-Charles Bazin, Ond\v{r}ej
  Jamri\v{s}ka, Alexander Sorkine-Hornung, and Daniel S\'{y}kora.
\newblock Painting by feature: Texture boundaries for example-based image
  creation.
\newblock {\em ACM Trans. Graph.}, 32(4):116:1--116:8, July 2013.

\bibitem{luo2012multi}
Linjie Luo, Hao Li, Sylvain Paris, Thibaut Weise, Mark Pauly, and Szymon
  Rusinkiewicz.
\newblock Multi-view hair capture using orientation fields.
\newblock In {\em Computer Vision and Pattern Recognition (CVPR), 2012 IEEE
  Conference on}, pages 1490--1497. IEEE, 2012.

\bibitem{mohammed2009visio}
Umar Mohammed, Simon~JD Prince, and Jan Kautz.
\newblock Visio-lization: generating novel facial images.
\newblock {\em ACM Transactions on Graphics (TOG)}, 28(3):57, 2009.

\bibitem{nguyen2008image}
Minh~Hoai Nguyen, Jean-Francois Lalonde, Alexei~A Efros, and Fernando De~la
  Torre.
\newblock Image-based shaving.
\newblock {\em Computer Graphics Forum}, 27(2):627--635, 2008.

\bibitem{olszewski2017realistic}
Kyle Olszewski, Zimo Li, Chao Yang, Yi Zhou, Ronald Yu, Zeng Huang, Sitao
  Xiang, Shunsuke Saito, Pushmeet Kohli, and Hao Li.
\newblock Realistic dynamic facial textures from a single image using gans.
\newblock In {\em IEEE International Conference on Computer Vision (ICCV)},
  pages 5429--5438, 2017.

\bibitem{park2019SPADE}
Taesung Park, Ming-Yu Liu, Ting-Chun Wang, and Jun-Yan Zhu.
\newblock Semantic image synthesis with spatially-adaptive normalization.
\newblock In {\em Proceedings of the IEEE Conference on Computer Vision and
  Pattern Recognition}, 2019.

\bibitem{perez2003poisson}
Patrick P{\'e}rez, Michel Gangnet, and Andrew Blake.
\newblock Poisson image editing.
\newblock {\em ACM Transactions on graphics (TOG)}, 22(3):313--318, 2003.

\bibitem{Portenier:2018}
Tiziano Portenier, Qiyang Hu, Attila Szab\'{o}, Siavash~Arjomand Bigdeli, Paolo
  Favaro, and Matthias Zwicker.
\newblock Faceshop: Deep sketch-based face image editing.
\newblock {\em ACM Trans. Graph.}, 37(4):99:1--99:13, July 2018.

\bibitem{DBLP:journals/corr/RonnebergerFB15}
Olaf Ronneberger, Philipp Fischer, and Thomas Brox.
\newblock U-net: Convolutional networks for biomedical image segmentation.
\newblock {\em CoRR}, abs/1505.04597, 2015.

\bibitem{RuderDB2016}
Manuel Ruder, Alexey Dosovitskiy, and Thomas Brox.
\newblock Artistic style transfer for videos.
\newblock Technical report, arXiv:1604.08610, 2016.

\bibitem{saito2016photorealistic}
Shunsuke Saito, Lingyu Wei, Liwen Hu, Koki Nagano, and Hao Li.
\newblock Photorealistic facial texture inference using deep neural networks.
\newblock {\em arXiv preprint arXiv:1612.00523}, 2016.

\bibitem{sangkloy2017scribbler}
Patsorn Sangkloy, Jingwan Lu, Chen Fang, Fisher Yu, and James Hays.
\newblock Scribbler: Controlling deep image synthesis with sketch and color.
\newblock In {\em IEEE Conference on Computer Vision and Pattern Recognition
  (CVPR)}, volume~2, 2017.

\bibitem{selim2016painting}
Ahmed Selim, Mohamed Elgharib, and Linda Doyle.
\newblock Painting style transfer for head portraits using convolutional neural
  networks.
\newblock {\em ACM Transactions on Graphics (ToG)}, 35(4):129, 2016.

\bibitem{shrivastava2017learning}
Ashish Shrivastava, Tomas Pfister, Oncel Tuzel, Joshua Susskind, Wenda Wang,
  and Russell Webb.
\newblock Learning from simulated and unsupervised images through adversarial
  training.
\newblock In {\em Proceedings of the IEEE Conference on Computer Vision and
  Pattern Recognition}, pages 2107--2116, 2017.

\bibitem{Simonyan14c}
K. Simonyan and A. Zisserman.
\newblock Very deep convolutional networks for large-scale image recognition.
\newblock {\em CoRR}, abs/1409.1556, 2014.

\bibitem{upchurch2016deep}
Paul Upchurch, Jacob Gardner, Kavita Bala, Robert Pless, Noah Snavely, and
  Kilian Weinberger.
\newblock Deep feature interpolation for image content changes.
\newblock In {\em Proc. IEEE Conf. Computer Vision and Pattern Recognition},
  pages 6090--6099, 2016.

\bibitem{ward2007survey}
Kelly Ward, Florence Bertails, Tae-Yong Kim, Stephen~R Marschner, Marie-Paule
  Cani, and Ming~C Lin.
\newblock A survey on hair modeling: Styling, simulation, and rendering.
\newblock {\em IEEE transactions on visualization and computer graphics},
  13(2):213--234, 2007.

\bibitem{wei2009state}
Li-Yi Wei, Sylvain Lefebvre, Vivek Kwatra, and Greg Turk.
\newblock State of the art in example-based texture synthesis.
\newblock In {\em Eurographics 2009, State of the Art Report, EG-STAR}, pages
  93--117. Eurographics Association, 2009.

\bibitem{Wei:2000:FTS}
Li-Yi Wei and Marc Levoy.
\newblock Fast texture synthesis using tree-structured vector quantization.
\newblock In {\em Proceedings of the 27th Annual Conference on Computer
  Graphics and Interactive Techniques}, SIGGRAPH '00, pages 479--488, 2000.

\bibitem{weng2013hair}
Yanlin Weng, Lvdi Wang, Xiao Li, Menglei Chai, and Kun Zhou.
\newblock Hair interpolation for portrait morphing.
\newblock {\em Computer Graphics Forum}, 32(7):79--84, 2013.

\bibitem{4069262}
Y. Wexler, E. Shechtman, and M. Irani.
\newblock Space-time completion of video.
\newblock {\em IEEE Transactions on Pattern Analysis and Machine Intelligence},
  29(3):463--476, March 2007.

\bibitem{Xing:2019:HID}
Jun Xing, Koki Nagano, Weikai Chen, Haotian Xu, Li-yi Wei, Yajie Zhao, Jingwan
  Lu, Byungmoon Kim, and Hao Li.
\newblock Hairbrush for immersive data-driven hair modeling.
\newblock In {\em Proceedings of the 32Nd Annual ACM Symposium on User
  Interface Software and Technology}, UIST '19, 2019.

\bibitem{yeh2017semantic}
Raymond~A Yeh, Chen Chen, Teck~Yian Lim, Alexander~G Schwing, Mark
  Hasegawa-Johnson, and Minh~N Do.
\newblock Semantic image inpainting with deep generative models.
\newblock In {\em Proceedings of the IEEE Conference on Computer Vision and
  Pattern Recognition}, pages 5485--5493, 2017.

\bibitem{yuksel2009hair}
Cem Yuksel, Scott Schaefer, and John Keyser.
\newblock Hair meshes.
\newblock {\em ACM Transactions on Graphics (TOG)}, 28(5):166, 2009.

\bibitem{zhang2017stackgan}
Han Zhang, Tao Xu, Hongsheng Li, Shaoting Zhang, Xiaogang Wang, Xiaolei Huang,
  and Dimitris~N Metaxas.
\newblock Stackgan: Text to photo-realistic image synthesis with stacked
  generative adversarial networks.
\newblock In {\em Proceedings of the IEEE International Conference on Computer
  Vision}, pages 5907--5915, 2017.

\bibitem{zhu2016generative}
Jun-Yan Zhu, Philipp Kr{\"a}henb{\"u}hl, Eli Shechtman, and Alexei~A Efros.
\newblock Generative visual manipulation on the natural image manifold.
\newblock In {\em European Conference on Computer Vision}, pages 597--613.
  Springer, 2016.

\bibitem{zhu2017unpaired}
Jun-Yan Zhu, Taesung Park, Phillip Isola, and Alexei~A Efros.
\newblock Unpaired image-to-image translation using cycle-consistent
  adversarial networks.
\newblock {\em arXiv preprint arXiv:1703.10593}, 2017.

\end{thebibliography}
